\documentclass{article}

\usepackage[preprint]{corl_2025} 
\usepackage{graphicx}
\usepackage{wrapfig}
\usepackage{eucal}
\usepackage{color}
\usepackage{amsmath}
\usepackage{lipsum} 
\usepackage[T1]{fontenc}
\usepackage{newtxtext}
\usepackage[cmintegrals]{newtxmath}
\usepackage{bm} 
\usepackage[linesnumbered,ruled,vlined]{algorithm2e}
\usepackage{subfigure}
\usepackage[utf8]{inputenc}
\usepackage{graphicx}
\usepackage{graphics}
\usepackage{siunitx}
\usepackage{multirow}
\usepackage{float}
\usepackage{caption}
\usepackage{gensymb}
\usepackage{hyperref}
\usepackage{epstopdf}
\usepackage{multicol,multirow}
\usepackage{booktabs}
\usepackage{textcomp}
\usepackage{xcolor}
\usepackage{wrapfig}
\usepackage{amsmath}
\usepackage{cleveref}

\usepackage{enumitem}

\graphicspath{{}}

\definecolor{red}{rgb}{1.00,0.00,0.00}
\definecolor{lightred}{rgb}{1.00,0.3,0.3}
\definecolor{blue}{rgb}{0.00,0.00,1.00}
\definecolor{green}{rgb}{0.1,0.50,0.1}
\definecolor{yellow}{rgb}{0.5,0.5,0.0}
\definecolor{white}{rgb}{1,1,1}
\definecolor{gray}{rgb}{0.6,0.6,0.6}
\newcommand{\cred}[1] {\textcolor{red}{#1}}

\usepackage{array, make cell, rotating}
\setcellgapes{2pt}
\newcommand\nocell[1]{\multicolumn{#1}{c|}{}}

\title{VITAL: Interactive Few-Shot Imitation Learning via Visual Human-in-the-Loop Corrections}

\author{
  Hamidreza Kasaei$^\clubsuit$, Mohammadreza Kasaei$^\spadesuit$\\
  $^\clubsuit$Department of Artificial Intelligence, 
  University of Groningen,
  The Netherlands\\
  $^\spadesuit$ School of Informatics, University
of Edinburgh, UK. \\
  \texttt{hamidreza.kasaei@rug.nl, m.kasaei@ed.ac.uk} \\
}

\begin{document}
\maketitle

\begin{abstract}
   Imitation Learning (IL) has emerged as a powerful approach in robotics, allowing robots to acquire new skills by mimicking human actions. Despite its potential, the data collection process for IL remains a significant challenge due to the logistical difficulties and high costs associated with obtaining high-quality demonstrations. To address these issues, we propose a large-scale data generation from a handful of demonstrations through data augmentation in simulation.
   Our approach leverages affordable hardware and visual processing techniques to collect demonstrations, which are then augmented to create extensive training datasets for imitation learning. By utilizing both real and simulated environments, along with human-in-the-loop corrections, we enhance the generalizability and robustness of the learned policies. We evaluated our method through several rounds of experiments in both simulated and real-robot settings, focusing on tasks of varying complexity, including bottle collecting, stacking objects, and hammering. Our experimental results validate the effectiveness of our approach in learning robust robot policies from simulated data, significantly improved by human-in-the-loop corrections and real-world data integration. Additionally, we demonstrate the framework's capability to generalize to new tasks, such as setting a drink tray, showcasing its adaptability and potential for handling a wide range of real-world manipulation tasks. A video of the experiments has been attached as a part of the supplementary materials. 
    
\end{abstract}



\section{Introduction}

Imitation Learning~\cite{schaal1999imitation} has emerged as a fundamental approach in robotics, enabling robots to acquire new skills by observing and mimicking human actions~\cite{jang2022bc,mandlekar2021matters}. This method holds significant promise for developing complex robotic behaviors without the need for explicitly programming the robot. However, the primary challenge with imitation learning lies in the data collection process~\cite{brohan2022rt, zhang2018deep,mandlekar2020learning,finn2017one}. On the one hand, acquiring high-quality demonstrations is costly and logistically challenging. On the other hand, the need for extensive and diverse demonstrations to train models magnifies these difficulties, as each new task or environment often requires a fresh set of demonstrations~\cite{wen2022you,valassakis2022demonstrate}. To address these challenges, teleoperation has been proposed as a viable solution. In recent years, there has been a growing interest in the development of teleoperation interfaces that facilitate the learning and execution of various household and industrial tasks~\cite{aldaco2024aloha,wang2024dexcap,fu2024mobile,schwarz2021nimbro}. 

One example of such a system is the ALOHA platform~\cite{aldaco2024aloha, fu2024mobile}, which has garnered considerable attention from the robotics research community. Such teleoperation platforms enable the execution of various tasks such as pushing chairs, using cabinets, wiping spills, and even cooking shrimp. Despite the advantages offered by such interfaces, there remain significant challenges that hinder their widespread adoption. One of the primary issues is that such teleoperation interfaces are hardware-specific and costly. The cost of such interfaces is approximately $\$30K$, which makes them non-scalable and expensive for many research laboratories and practical applications. To overcome these challenges, we propose a low-cost visual teleoperation interface to collect high-quality demonstrations for long-horizon manipulation tasks.

In robot manipulation tasks, IL is often trained via in-domain demonstrations, which correspond to the data obtained directly from the deployment environment~\cite{wang2023mimicplay}. Although it is widely accepted that gathering direct examples from the actual robot is the most efficient way to learn long-horizon manipulation tasks, we show that collecting human demonstrations in real and digital twin environments can produce better outcomes for real-world tasks. Our approach allows the creation of a large dataset of augmented demonstrations based on a few initial demonstrations. By leveraging the flexibility and functionalities of our digital twin environment, we can enhance the number of demonstrations and the generalizability of the learned policy. These augmented demonstrations are then used to train a policy, which is fine-tuned using real-world demonstrations. Experimental results show that our approach significantly improves the performance of the robot in real-world scenarios. In summary, the main contributions of this work include:

\begin{itemize}[leftmargin=0.1in]
\item Large-scale data generation from a handful of demonstrations through data augmentation in simulation.

    \item Incorporate human-in-the-loop corrections to refine and improve the robot's performance in complex tasks. This method allows for interactive learning and adaptation, ensuring higher success rates in real-world applications.
    \item Successful deployment of policies trained on such data to the real robot with only a handful of demonstrations.

\end{itemize}

\section{Related Work}
\textbf{Robot Learning with Teleoperation:} IL is a powerful technique for robot learning, enabling robots to acquire new skills by mimicking human demonstrations~\cite{arents2022smart,ravichandar2020recent,johns2021coarse-to-fine}. However, collecting in-domain demonstrations is often difficult and not scalable due to the need for physical interaction with the robot hardware. This process is also expensive as it requires constant access to the robot. Some works use scripted ``\textit{expert}'' in simulation to generate demonstrations, but scaling these approaches to more complex tasks is challenging~\cite{zheng2022imitation,mandlekar2021matters}. Other IL methods leveraged visual deep neural networks to train reactive models from images captured during demonstrations~\cite{florence2019self,florence2022implicit,wang2024eve}. While these models provide greater flexibility, they necessitate collecting a large number of demonstrations to learn simple pick-and-place tasks, making them user and computationally expensive~\cite{wang2023mimicplay}. Other approaches utilize human activity videos to guide robot learning but often fall short in training 6-DoF manipulation policies~\cite{mandikal2022dexvip, sivakumar2022robotic}. These methods typically require in-domain teleoperation data to bridge the gap between video demonstrations and robotic execution~\cite{wang2024dexcap}.
A common problem with these IL methods is the lack of corrective feedback, leading to the accumulation of errors over the length of the demonstration, known as the \textit{compounding errors problem}~\cite{shi2023waypointbased}. Similar to our approach, some methods have focused on dividing demonstrations into several waypoints to address this issue~\cite{shi2023waypointbased,james2022q,akgun2012keyframe}. Unlike these approaches, we aim to automatically extract the waypoints of a demonstration without adding any burden on the human user. Other IL methods used  teleoperation interfaces, such as ALOHA~\cite{aldaco2024aloha}, for mapping joints' value from master to slave robots. However, these approaches are quite expensive and not scalable as they require four robots for  manipulation: two masters and two slaves. In contrast, we propose a low-cost teleoperation interface that allows controlling robot end-effectors to perform complex  tasks and collect demonstrations in both real and simulation environments. Teleoperation within a digital twin of the robot enables us to scale up the training data easily without constant access to the real robot. Additionally, we introduce a human-in-the-loop correction mechanism to repair faulty demonstrations, enhancing the robot's performance in complex tasks with minimal effort. 

{To contrast with existing methods, our approach differs significantly from both DAgger and prior human-in-the-loop teleoperation systems~\cite{kelly2019hg,mandlekar2020human}. Unlike DAgger\cite{kelly2019hg}, which relies on offline expert relabeling of trajectories collected by the policy, we propose a real-time correction mechanism where humans intervene only when necessary, either through fine-grained 3D wrist delta adjustments or full teleoperation. This eliminates the high annotation burden and enables scalable policy refinement. Compared to previous human-in-the-loop frameworks such as Mandlekar et al.~\cite{mandlekar2020human}, which rely on smartphone-based remote interventions and emphasize bottleneck traversal through weighted regression, our method takes a more general and robust approach to policy correction. While prioritizing bottlenecks may introduce bias by assuming certain regions of a task are more critical than others, our framework avoids such assumptions by responding to errors wherever they occur.} 

\textbf{Sim2Real Transfer:} Some approaches, such as MimicGen~\cite{mandlekar2023mimicgen} and CyberDemo~\cite{wang2024cyberdemo}, attempt to collect demonstrations in simulation to reduce the cost of collecting demonstrations on robots. However, due to the sim-to-real gap, deploying the learned policy on a real-world robot remains challenging. Other methods use VR for teleoperation and data collection~\cite{brandfonbrener2023visual,hetrick2020comparing}, such as DIME~\cite{arunachalam2023dexterous}. Qin et al.~\cite{schwarzer2021pretraining} and Wang et al.~\cite{wang2024cyberdemo} proposed tracking the user's figures through an external camera for teleoperation. Approaches like DexMV~\cite{qin2022dexmv}, DexVIP~\cite{mandikal2022dexvip}, and VideoDex~\cite{shaw2023videodex} use everyday videos to learn motion priors. Due to the domain gap, these methods are not directly applicable to robots and often require another round of training.
In contrast, we collect demonstrations in the digital twin of the robot and develop data augmentation techniques at the trajectory level rather than the image data level. Most of the teleoperation interfaces have been designed either for simulation~\cite{mandlekar2023mimicgen} or for real robot ~\cite{aldaco2024aloha} setups. Unlike these approaches, we proposed a low-cost  teleoperation interface that can be used to collect demonstrations using real-robot and its simulation. We showed that policies trained on a mix of simulated and real demonstrations achieve better results than those trained solely on simulation or real-robot data. 
\vspace{-2mm}
\section {Method}
\subsection{Collecting Demonstrations through Teleoperation}
\begin{wrapfigure}{r}{0.4\textwidth}
\vspace{-3mm}
    \centering \includegraphics[width=0.4\textwidth]{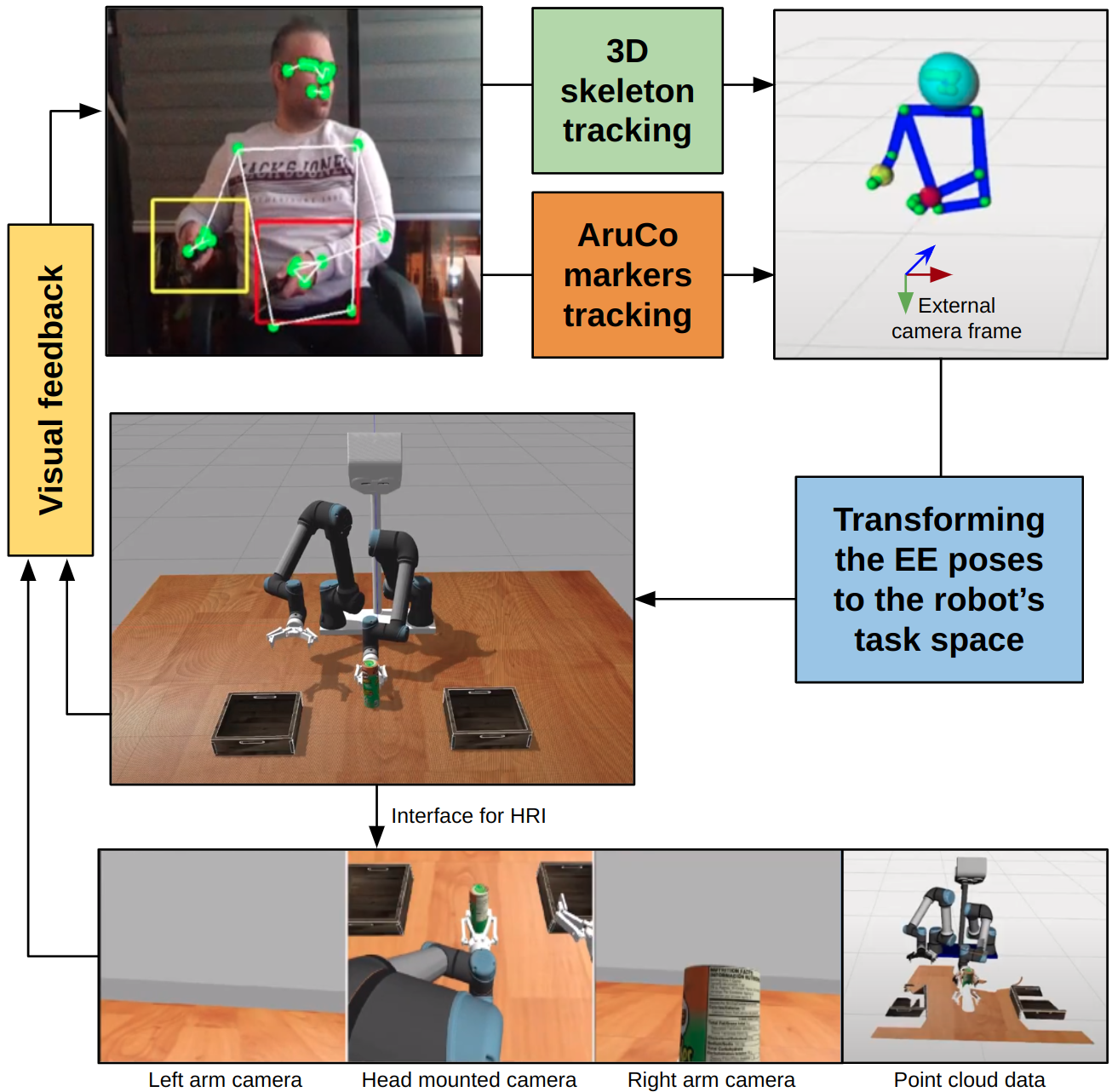}
    \caption{Overview of the proposed low-cost teleportation interface.}
    \label{fig:overview}
    \vspace{-3mm}
\end{wrapfigure}
    
The architecture of our low-cost visual teleoperation system is illustrated in Figure~\ref{fig:overview}. 
Given the differences in body morphology and arm length between humans and the robot, two scaling factors, $\alpha$ for the right arm and $\beta$ for the left arm,  are applied to bridge the gap, thereby making teleoperation more intuitive and precise. 
For controlling the grippers of the robot's arms, we utilized two Bluetooth selfie sticks. These sticks were adapted to send commands for opening and closing the grippers. 
To track the sticks' orientation, we attached AruCo markers~\citep{Aruco2014, GarridoJurado2015} to the sticks.
To efficiently manage the complexity of long-horizon tasks and reduce the compounding errors, we employ task decomposition. Each task, $T$, is decomposed into a sequence of subtasks, $S_n$, $T = (S_1, S_2, \dots, S_n)$. The start and end poses of the trajectory for each subtask are defined based on the current pose of the manipulator and the pose of the object that should be grasped or placed. 
The detection of these subtasks is facilitated by the commands given by the user for grasping or placing objects. When a user issues a command, the system identifies the relevant subtask, captures the necessary pose information, and creates a trajectory segment for that subtask. By structuring tasks this way, we ensure that the demonstration data is organized and can be effectively used for training purposes.

To replicate the physical environment in the simulation, we build a digital twin of our robot in Gazebo. This includes accurately modeling the robot and objects to ensure consistency between the simulated and real-world environments. During the demonstration phase, we performed a task in Gazebo while recording all relevant ROS topics, including the initial state of the objects, joint values, end-effector poses, outputs from RGB-D cameras (head-mounted and on the left and right arms), commands from the instructor, activity videos with corresponding 3D skeleton data, object poses, and success or failure indicators, at their respective publication rates into a ROS bag file. The comprehensive nature of this recorded data ensures that all necessary information is captured for the next stage of the methodology, where data augmentation will be applied to enhance the dataset.

\subsection {Augmenting Demonstration in Simulation}

To enhance the collected demonstration and improve the robustness of the learned policies, we perform data augmentation on the trajectory level (i.e. subtasks). Our augmentation method uses several techniques to generate a diverse set of trajectories from the initial demonstrations. First, we extract waypoints from the recorded bag file by reading the positions and orientations of the robot's end effectors. These waypoints are then used to fit a polynomial trajectory. We exclude the last few points from the polynomial fitting to ensure the end of the trajectory which represents the affordance pattern is preserved. We then apply uniform sampling to the trajectory by calculating the cumulative arc length of the waypoints to ensure uniform spacing between the sampled points. 
This method helps in creating a smooth and evenly distributed trajectory, which is then segmented into smaller parts for detailed analysis and augmentation.

We first augment the trajectory by adding Gaussian noise to the waypoints. This involves introducing random variations to the trajectory points to simulate real-world imperfections and variations. By adding noise, we can create multiple versions of the trajectory, enhancing the model's ability to generalize across different scenarios. The noise can be applied to all waypoints or selectively to preserve certain critical points, such as the trajectory's endpoint.
We also perform translations and flips on the trajectory. Translating the waypoints involves shifting the entire trajectory in the x, y, and z directions, which helps in creating varied instances of the task. Flipping the trajectory along the x-axis generates a mirrored version, providing additional diversity in the dataset and making the initial trajectory of one arm useful for the other arm. Furthermore, we sample new initial points within specified bounds and connect these points to the nearest waypoints in the existing trajectory. This process involves finding the nearest index in the trajectory and creating new paths from these initial points to the rest of the trajectory. By varying the starting points, we can simulate different initial conditions for the task. Each trajectory is annotated with an arm identifier, distinguishing between left and right arm movements. This comprehensive augmentation process significantly expands the dataset, making it thousands of times larger than the original set of demonstrations. However, augmented trajectories can sometimes be imperfect and not executable due to the robot's arm kinematics, potentially leading to task failures. To ensure the reliability of the augmented data, we validate each augmented trajectory in the digital twin environment. Only successfully executed trajectories are used for training a policy. This way we ensure that the learned policies are robust and capable of handling a wide range of real-world conditions.

\subsection {Hierarchical Policy Learning}


To effectively learn and execute long-horizon tasks, we formulate the learning policy as a hierarchical policy learning problem through Behavioral Cloning (BC). This approach involves training a high-level policy to select a sequence of subtasks and low-level policies to execute each subtask. 
For a given task, we assume that we can detect the sequence of subtasks based on the demonstration by detecting commands for grasp or release (indicated by pressing the bottom of the selfie stick). These commands allow us to identify the start and end poses for each subtask, facilitating the decomposition of long-horizon tasks into a series of subtasks.

The high-level policy, $\pi_H$, is responsible for selecting the appropriate subtask, $S_i$, based on the current state of the environment and the task progression. This high-level policy can be implemented as a state machine or a learned model. In this work, we use a state machine that receives the current state of the robot, and the pose of the objects to infer the current subtask, $\pi_H(\textbf{o}, r)\rightarrow S_i$, where $\textbf{o}$ shows the current state of objects and $r$ is the state of the robot. Alternatively, a learned model could receive the task ID and the state of the objects and robot, and predict the subtask.


For the low-level policy, we propose a unified multi-subtask policy learning, $\pi_{MS}$, that can handle all subtasks. This policy receives the predicted subtask ID, $S_i$, along with the start, $p_s$, and end poses, $p_e$, and generates the appropriate trajectory conditioned on the input subtask, $\pi_{MS}(S_i,p_s, p_e) \rightarrow$ \textit{trajectory}. The training process begins by loading the augmented trajectory data of a task. The trajectories are padded to ensure uniform length, which is critical for batch processing in neural networks. Our model is designed to predict the intermediate points of a trajectory given the start and end points. The high-level policy coordinates these subtasks, ensuring smooth transitions and overall task completion.

\subsection {Residual Learning with Human-in-the-Loop}

\begin{wrapfigure}{r}{0.5\textwidth}
    \vspace{-5mm}
\includegraphics[width=0.5\textwidth]{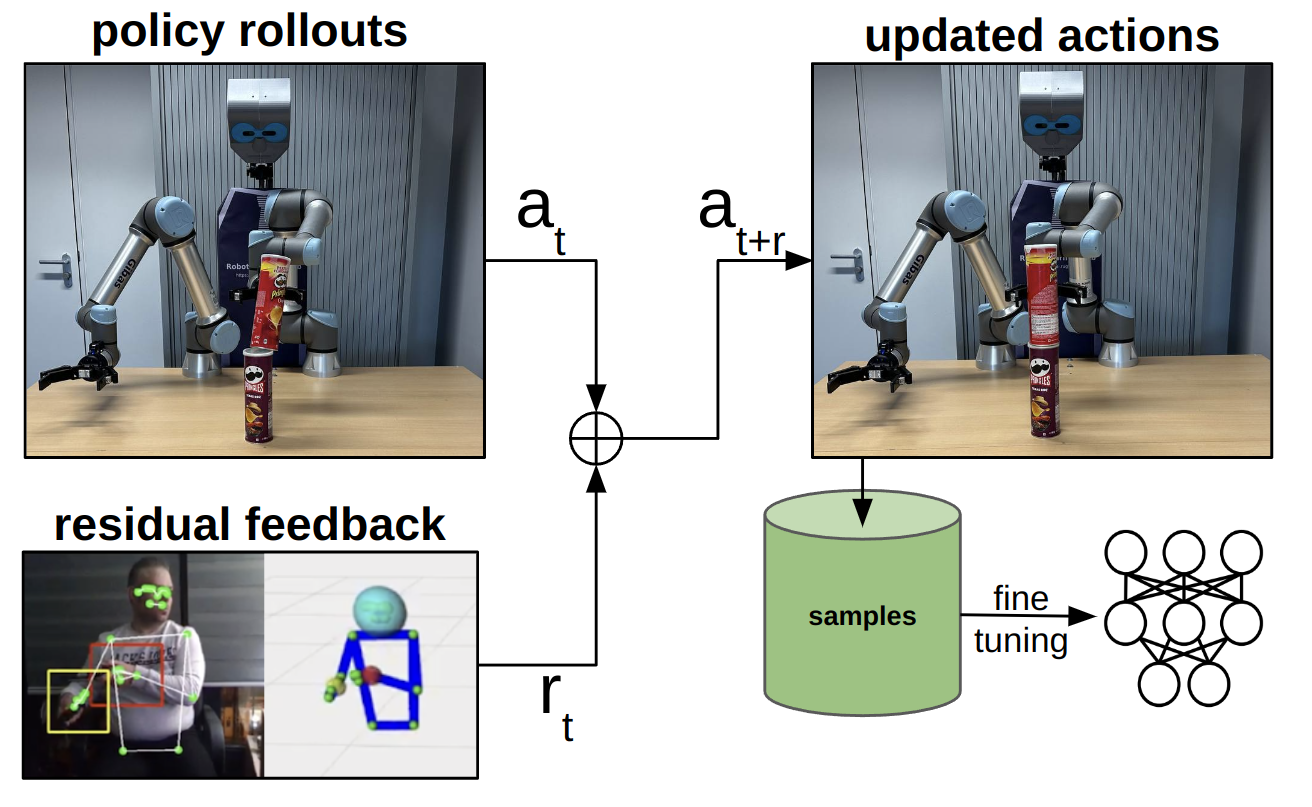}
    \caption{Overview of the residual learning based on human-in-the-loop feedback.}
    \label{fig:all_objs}
    \vspace{-3mm}
\end{wrapfigure}
Despite the robustness of the trained policy, there can be scenarios where the robot fails to complete the task due to a sim-to-real gap, pose estimation errors, or small errors in executing a trajectory. To address these issues, we incorporate a residual learning approach with human-in-the-loop corrections during the policy rollouts. When the robot encounters a failure during task execution, a human operator steps in to correct the errors. The user plays the recorded rosbag file of the experiment and provides the correction feedback whenever it is needed. The correction involves measuring the 3D delta position changes of the wrist of the left and right hands of the demonstrator relative to the initial poses of the hands. These delta changes are then scaled by empirically determined factors $\alpha$ (for the right hand) and $\beta$ (for the left hand), both $< 0.1$ to avoid fast movements, and added to the robot's wrist movements. This minimal adjustment helps the robot to accomplish the task with improved accuracy. If the user observes a large error in the current prediction of the policy, they can switch the system to full teleoperation mode. In this mode, the robot will directly follow the human wrist movements to accomplish the task. 
The data from these human-in-the-loop corrections is recorded into a separate dataset, 
$D'$. This dataset includes both residual corrections and instances where full teleoperation was required. To fine-tune the policy, training data is sampled equally from both the original dataset $D$ and the new dataset $D'$. This balanced sampling ensures that the model learns from the augmented experiences and corrections provided by the human operator, thereby improving its performance and adaptability. This iterative process of learning ensures that the policy continuously improves and adapts to real-world conditions.

\section{Experimental Setup}
\vspace{-3mm}

We performed multiple rounds of experiments on both simulation and real-robot settings to validate our method. Specifically, we study the following questions: \textbf{(\textit{Q1})} \textit{Is it feasible to train robot policies using exclusively simulation data without relying on on-robot data? }
\textbf{\textit{(Q2)}} \textit{Which model architectures and data composition ratios are most effective for policy training? Specifically, what is the optimal balance between simulated and real-world data?}
\textbf{\textit{(Q3)}} \textit{To what extent does human-in-the-loop correction enhance policy performance when teleoperation data is limited?}
\textbf{\textit{(Q4)}} \textit{Can our framework successfully execute another complex  task, such as setting a drink tray, which is somewhat similar to one of the learned tasks?}

\subsection{Setup} 
\begin{wrapfigure}{r}{0.38\textwidth}
\vspace{-3mm}
    \centering    \includegraphics[width=0.38\textwidth]{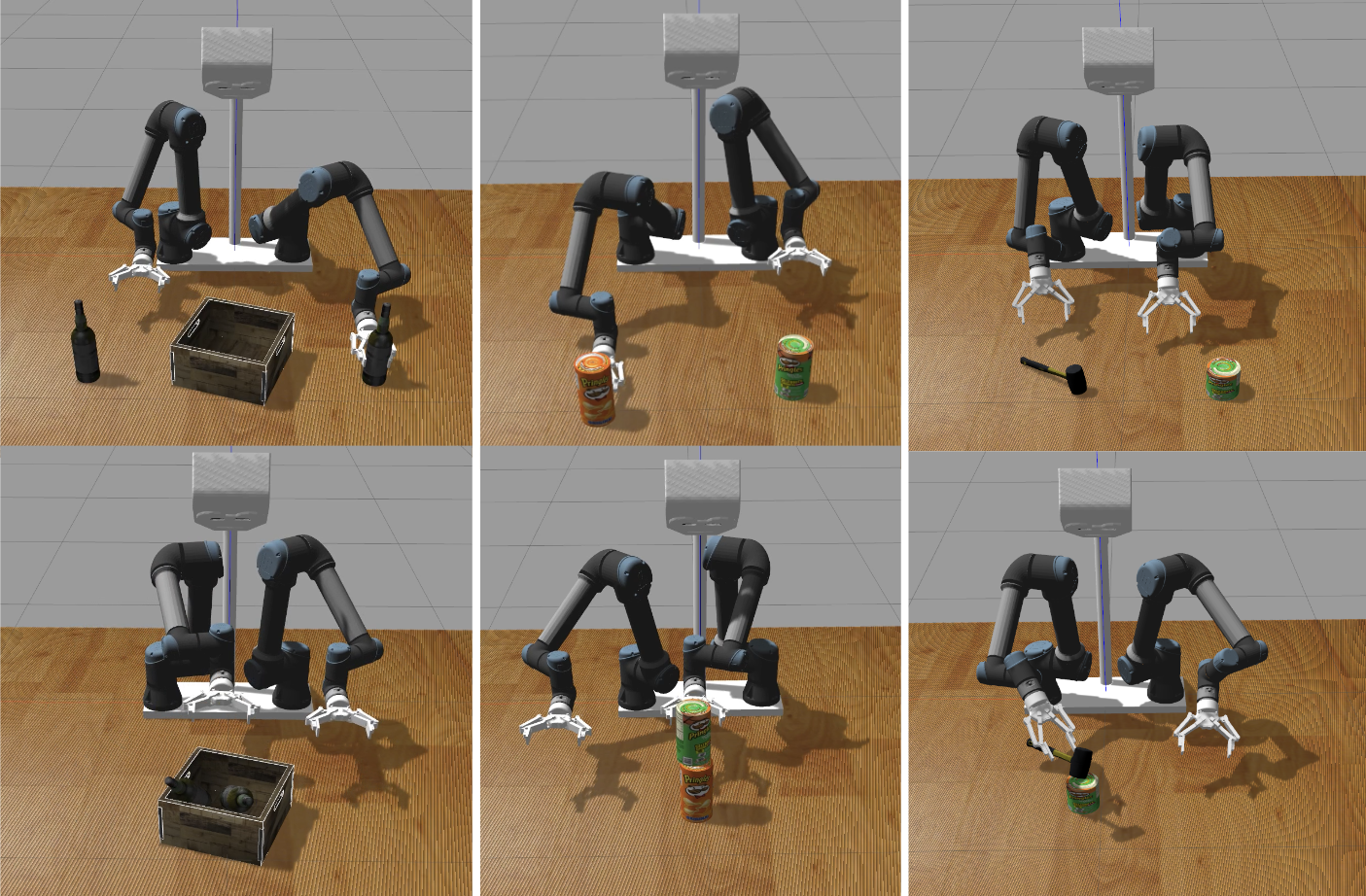}
    \caption{Three long-horizon evaluation tasks are used. (\textit{left}) Bottle Collecting; (\textit{center}) Stacking Pringles; (\textit{right}) Hammering.}
    \vspace{-3mm}
    \label{fig:tasks}
\end{wrapfigure}

\begin{wrapfigure}{r}{0.38\textwidth}
\vspace{-3mm}
    \centering    \includegraphics[width=0.38\textwidth]{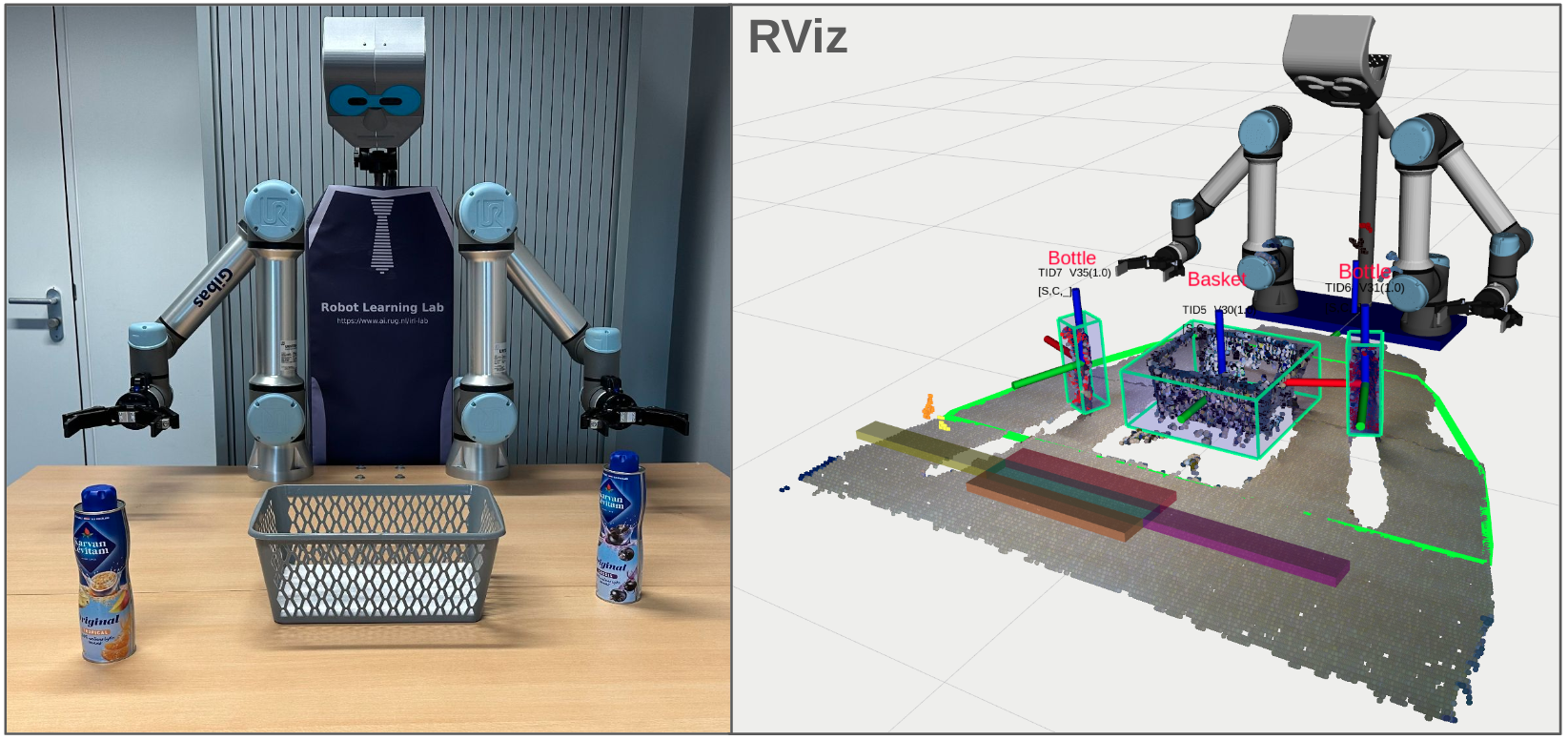}
    \caption{(\textit{left}) Our real dual-arm robot setup; (\textit{right}) Output of our perception including the bounding box, pose, and label of the objects.}
    \vspace{-3mm}
    \label{fig:real-robot}
\end{wrapfigure}

Our experimental setups, both in simulation and on real robots, are illustrated in Fig.~\ref{fig:tasks} and Fig.~\ref{fig:real-robot}, respectively. We developed a simulation environment in Gazebo, leveraging the ODE physics engine to closely mimic the behavior of our dual-arm robot. The hardware setup in both simulation and real robot includes an Asus Xtion camera, two Universal Robots (UR5e) equipped with Robotiq 2F-140 grippers, and an interface for managing the initiation and conclusion of experiments. For the Teleoperation interface, we use a RealSense D435 RGB-D camera to capture the motion of the user. All evaluations were performed with a PC running Ubuntu $18.04$ with a $3.20$ GHz Intel Xeon(R) $i7$, and a Quadro P$5000$ NVIDIA. 

\textbf{Tasks:} We designed three tasks of varying difficulty levels to test the performance of the teleoperation and the robustness of the learned policies: (\textit{i}) \textbf{\textit{Bottle Collecting}}: In this task, the robot should collect multiple bottles with different shapes from different locations. (\textit{ii}) \textbf{\textit{Stacking Pringles}}: In this task, the robot must stack Pringles cans on top of each other in various locations. This requires precise control and coordination to ensure the cans are aligned and stable. 
(\textit{iii}) \textbf{\textit{Hammering:}} The most challenging task involves the robot using a hammer to strike a cylinder. This task tests the robot's ability to handle a tool effectively. To evaluate the performance of the robot on each task, we performed 100 simulation and 10 real-robot experiments. In the case of real robot experiments, we used a 3D perception system to get the pose and label of the objects~\cite{kasaei2024lifelong,kasaei2024simultaneous}.


\textbf{Baselines and Metrics:} To determine the best model architecture for  manipulation tasks, we evaluated multiple Behavioral Cloning Recurrent Neural Network (BC-RNN) baselines. The baselines include: LSTM (Long Short-Term Memory), GRU (Gated Recurrent Unit), and Transformer-based network. PyTorch is used as a deep learning framework. We use Mean Squared Error (MSE) as the loss function for trajectory prediction. An Adam optimizer is employed to adjust the model weights during training. Early stopping is implemented to halt training if the validation loss does not improve for a specified number of epochs. Each network has been trained for 200 epochs. We split the data into training (70\%), validation (15\%), and test sets (15\%) to evaluate the model's performance. Details of the networks are discussed as a part of supplementary materials. The \textit{Task Success Rate} is used as the primary metric for evaluation. This metric is defined as the percentage of successfully completed tasks out of the total number of attempts. 

\vspace{-2mm}
\subsection{Results}
\begin{wrapfigure}{r}{0.38\textwidth}
\vspace{-5mm}
 \includegraphics[width=0.38\textwidth]{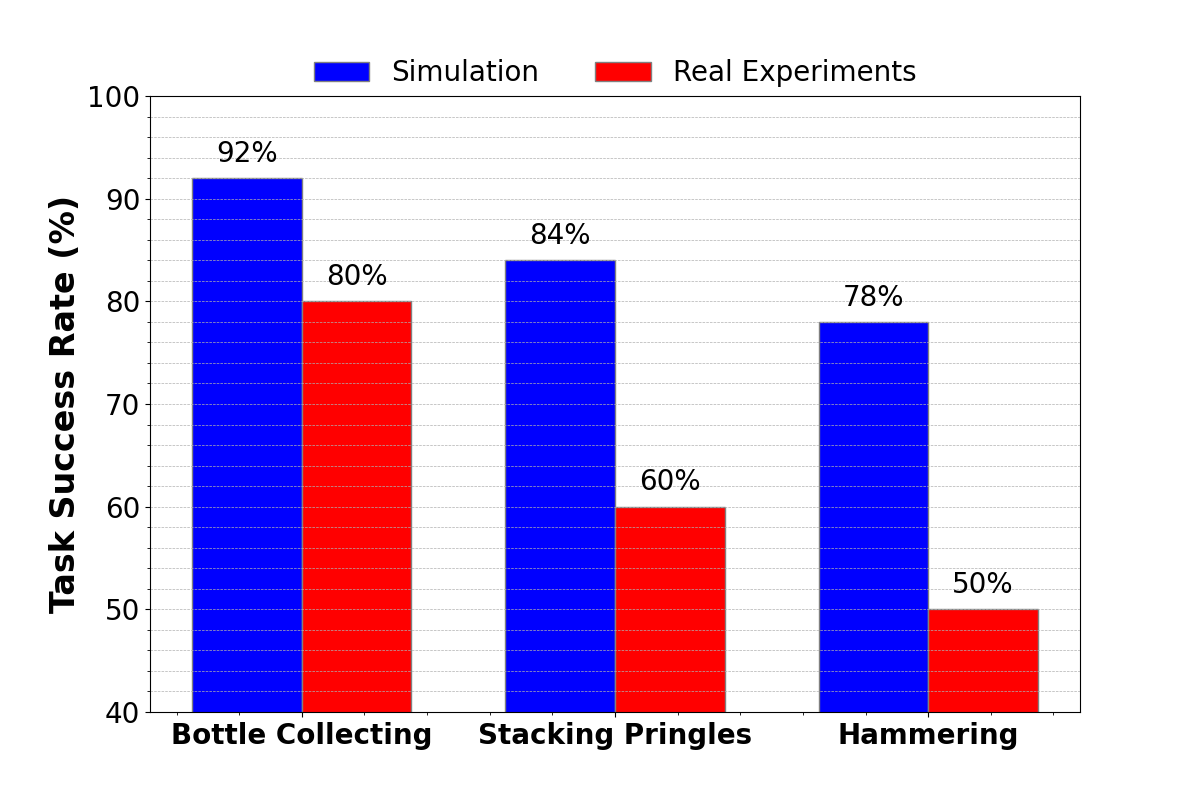}
    \caption{Results of Q1 evaluation.}
    \vspace{-5mm}
    \label{fig:Q1}
\end{wrapfigure}

\vspace{-3mm}
\textbf{Q1 - Feasibility of Training Policies with Simulation Data:}
We conducted extensive sets of experiments across all tasks by training the policy solely on augmented simulation trajectories. For each task, we collected five successful in-domain demonstrations and augmented them to 80K demonstrations. The BC-LSTM policy trained on this augmented data was then tested on both simulated and real robots. Results are summarized in Fig.~\ref{fig:Q1}. In simulation, the robot successfully performed most tasks. However, in real-world experiments, 
certain discrepancies such as minor inaccuracies in object grasping and placement were observed. Specifically, in the bottle collecting task, the robot achieved an average task success rate of 92\% in simulation and 80\% in real-robot scenarios. In the stacking objects task, the robot successfully completed 84\% of the experiments in simulation but only 60\% in real-world tests. Similarly, for the hammering task, the success rate was 78\% in simulation and 50\% in real-world experiments. The main failure in bottle collecting was due to inaccurate trajectory prediction, which led to inefficiencies in approaching, grasping, or placing the bottles. The primary reasons for failures in the stacking and hammering tasks were inaccuracies in object grasping. Specifically, when the robot approached to grasp the object, the tip of the gripper touched the object, causing it to move slightly. This offset often led to failures as objects falling during the last step of stacking or missing the target when hammering. To address these challenges, we will check the impact of the model architecture and the effect of incorporating real demonstrations in addition to digital-twin demonstrations for training the policy to address sim-to-real gap in the following subsection (\textbf{Q2}). Furthermore, we believe incorporating feedback mechanisms through vision sensing could provide information on the exact part of the object being grasped. This feedback would enable the system to update the trajectory accordingly by learning suitable residual corrections, thereby improving the accuracy and reliability of the task execution.

\begin{wrapfigure}{r}{0.38\textwidth}
\vspace{-4mm}
    \includegraphics[width=0.38\textwidth]{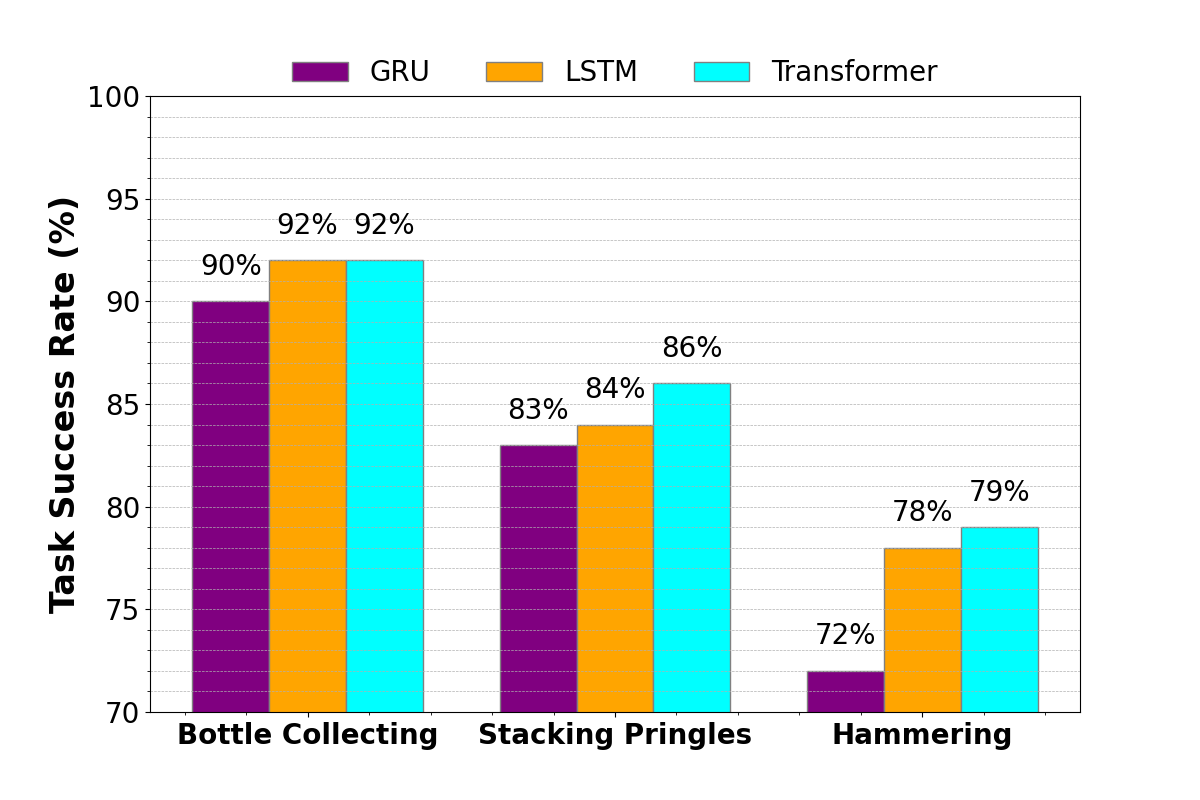}
    \vspace{-4mm}
    \caption{Results of Q2 evaluation.}
    \vspace{-3mm}
    \label{fig:Q2}
\end{wrapfigure}

\begin{figure}[!t]
    \centering  
    \includegraphics[width=\textwidth]{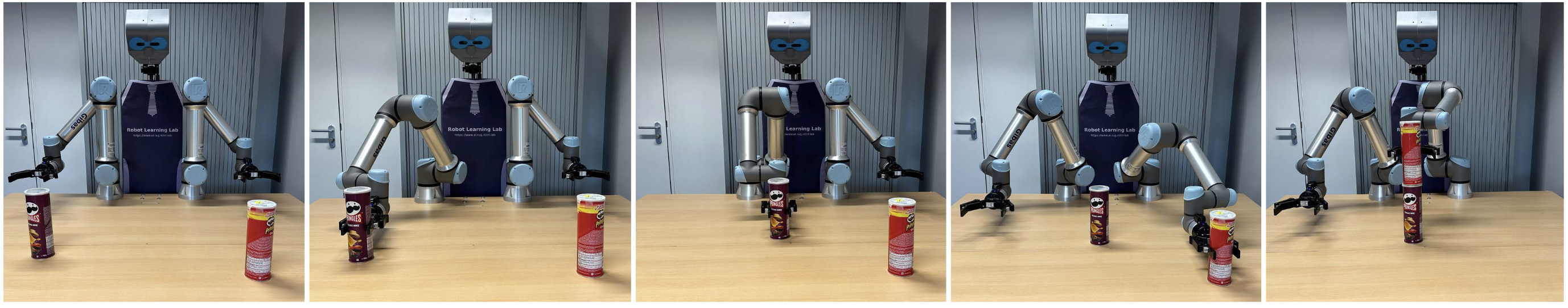}
    \caption{Sequence of snapshots demonstrating the stacking objects task performed by the real robot. 
    }
    \vspace{-2mm}
    \label{fig:stacking_task_real_robot}
\end{figure}

\textbf{Q2 - Impact of Model Architecture and Data Composition:} To understand the influence of different model architectures and the ratio of simulated to real data, we conducted experiments with various configurations. We first trained LSTM, GRU, and Transformer-based models solely on augmented simulation demonstrations and compared their performance across the proposed tasks in the simulation environment. After identifying the best-performing architecture, we trained policies using different ratios of simulated to real-robot demonstrations, including $70\%-30\%$, $50\%-50\%$, $30\%-70\%$, and $0-100\%$ mixes. For each task, we recorded five real-world demonstrations and five simulation demonstrations, which were then augmented to create a larger dataset. The final training data was sampled from these augmented demonstrations according to the specified ratios. This approach allowed us to determine the optimal balance between simulation and real-world data for training robust and effective robot policies.
The LSTM model provided a good balance of performance and training efficiency, whereas GRU models exhibited slightly lower performance but benefited from reduced computational overhead. Transformer-based models demonstrated considerable potential for handling more complex tasks; however, they demanded significantly more computational resources (see Fig.~\ref{fig:Q2}). 
Upon comparing the results, we identified that the optimal ratio of simulated to real data was 70\% simulated and 30\% real. This ratio offered the best generalization and performance in real-world tasks, enhancing the task success rate across all evaluated tasks. Specifically, for bottle collecting, the task success rate increased from 80\% to 90\%; for stacking Pringles, the success rate rose from 60\% to 80\% (see Fig.~\ref{fig:stacking_task_real_robot}); and for hammering, the success rate improved from 50\% to 70\%. This balanced approach effectively mitigated the sim-to-real gap, thereby enhancing the robustness and reliability of the learned policies.

\textbf{Q3: Effectiveness of Human-in-the-Loop Corrections:}
In this round of experiments, we investigated the effectiveness of human-in-the-loop correction in scenarios when the policy failed during task execution. We recorded 5 failure experiments for each task in the real robot. 
We then replayed the bag file and allowed the human user to correct the robot’s movements using residual learning or full teleoperation mode. The updated demonstrations were tested to ensure they could successfully accomplish the task. These corrected demonstrations were then augmented,  
\begin{wrapfigure}{r}{0.38\textwidth}
\vspace{-5mm}
    \includegraphics[width=0.38\textwidth]{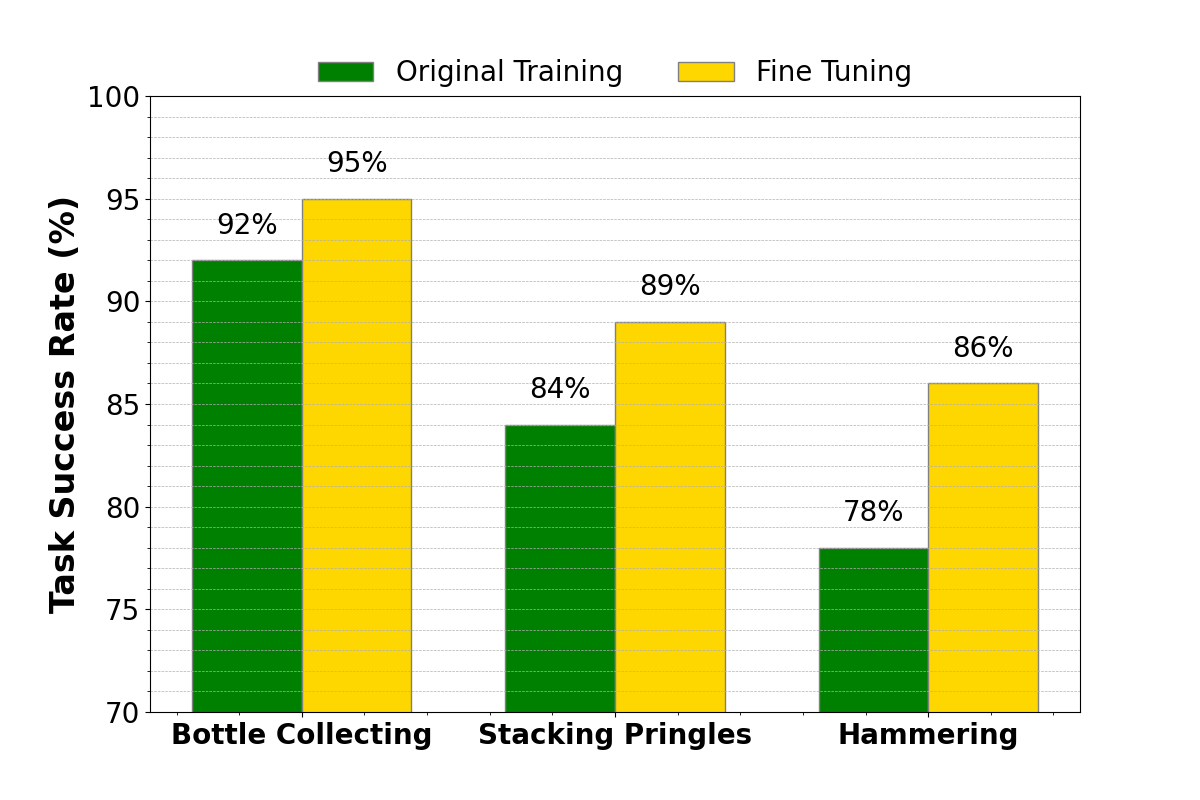}
     \vspace{-5mm}
    \caption{Results of Q3 evaluation.}
    \vspace{-5mm}
    \label{fig:Q3}
\end{wrapfigure}
and used for policy \textit{fine-tuning}. By comparing the results, we observed that human-in-the-loop corrections significantly improved task success rates, particularly for more complex tasks such as stacking and hammering. 
The inclusion of human corrections from dataset $D'$, resulted in an average improvement of approximately 5\% in task success rates, demonstrating the value of interactive learning. Specifically, the success rate for the hammering task increased from 78\% to 86\%, for the stacking task from 84\% to 89\%, and for the bottle collecting task from 92\% to 95\%. These results indicate that the policy was effectively learning from the corrections, as the number of correction feedback by the human user decreased over time. 

\textbf{Q4: Capability to Execute A New Complex  Task: Setting a Drink Tray}
To evaluate the generalization capabilities of our framework, we tested its ability to execute a new task that shares similarities with the learned tasks: setting a drink tray. 
This task involves placing a cup and a bottle on specific markers on a tray. The tray has two circular markers: one red for placing the cup and one blue for placing the bottle. The robot must pick up a cup and a bottle from a starting position and place them accurately on the designated red and blue markers on the tray. The policy was initially trained on the previously learned object stacking task. The trained policy was then evaluated only in simulation, as we could obtain precise positions of the markers and objects. The task was considered a success if the robot accurately placed the cup and bottle on the respective markers and a failure if the distance between the object and the marker exceeded 3 cm. We repeated this experiment 
100 times in simulation and 10 times in real, with the objects placed in random positions each time. The robot achieved a task success rate of $87\%$ in simulation and $80\%$ in the real robot. The primary source of error in the failure cases was slight misalignment during grasping, which led to minor placement inaccuracies, despite accurate trajectory predictions. Our approach successfully generalized to the new task, demonstrating that the learned policies could be adapted to similar but distinct tasks with minimal effort. This showcases the robustness and adaptability of our teleoperation and learning framework in handling a variety of real-world scenarios. We believe that incorporating feedback mechanisms through vision, such as detecting the exact positions of the markers and adjusting the trajectory accordingly, could further enhance the accuracy and reliability of task execution.

\vspace{-0mm}
\section{Conclusion and Limitations}
\label{sec:conclusion}
In this work, we presented a low-cost visual teleoperation system for manipulation tasks. Our approach leverages an RGB-D camera and visual processing techniques to collect demonstration data, which is then augmented and used to train robot policies. We employed a combination of simulated and real-world data, augmented with human-in-the-loop corrections, to improve the accuracy and reliability of the learned policies. Our system was evaluated on three tasks with differing complexity levels, demonstrating its effectiveness in both simulated and real-robot environments. The experimental results validate the effectiveness of our approach in learning robust policies from simulated data, enhanced by human-in-the-loop corrections, and real-world data integration. The framework successfully generalized to a new, complex task —setting a drink tray—demonstrating the adaptability of the learned policies to similar tasks with minimal effort. 

The main limitation of this work is that for tasks requiring a high degree of precision at certain steps (e.g., stacking, hammering, etc), waypoint trajectory learning may not be sufficient, and (closed-loop) vision feedback should be considered. To address this limitation, we plan to explore the integration of visual feedback mechanisms to further enhance the accuracy and reliability of task execution. By incorporating real-time visual feedback, the system can adjust and refine its actions based on the visual context, leading to more precise and robust manipulation capabilities.  Moreover, although we propose HITL correction as a means of addressing sim-to-real imperfections, its evaluation was limited to a small set of controlled interventions. We plan to explore more dynamic, on-the-fly HITL strategies and evaluate their impact over a broader range of tasks to fully quantify their contribution.

\clearpage


\bibliography{refs}  


\clearpage
\begin{center}
    \centering
    \Large
    \textbf{--- Supplementary Material ---\\
    VITAL: Interactive Few-Shot Imitation Learning via Visual Human-in-the-Loop Corrections}
\end{center}
    
\vspace{4 cm}

\setcounter{section}{0}

\section{Impact of Model Architecture and Data Composition (Q2)}
\begin{wraptable}{r}{0.5\textwidth}
\vspace{-4mm}
\centering
\caption{Hyperparameters for GRU, LSTM, and Transformer Models}
\resizebox{\linewidth}{!}{
\begin{tabular}{|l|c|c|c|}
\hline
\textbf{Hyperparameter}       & \textbf{GRU}   & \textbf{LSTM}  & \textbf{Transformer} \\ \hline
Epochs                        & 100            & 100            & 100                  \\ \hline
Batch size                    & 32             & 32             & 32                   \\ \hline
Learning rate                 & 0.001          & 0.001          & 0.001                \\ \hline
Loss function                 & MSE            & MSE            & MSE                  \\ \hline
Early stopping patience       & 10             & 10             & 10                   \\ \hline
Input size (first layer)      & 15             & 15            & 15                    \\ \hline
Hidden size (GRU/LSTM layers) & 50             & 50             & N/A                  \\ \hline
Number of GRU/LSTM layers     & 2              & 2              & N/A                  \\ \hline
Number of encoder layers      & N/A            & N/A            & 2                    \\ \hline
Number of decoder layers      & N/A            & N/A            & 2                    \\ \hline
Number of heads               & N/A            & N/A            & 4                    \\ \hline
d\_model                      & N/A            & N/A            & 128                  \\ \hline
\end{tabular}}
\label{tab:params}
\vspace{-6mm}
\end{wraptable}
\subsection{Network architectures and hyperparameters}
Table~\ref{tab:params} summarizes the hyperparameters used for the GRU, LSTM, and Transformer models in our experiments. While certain features are common across all three models, each model also has specific characteristics that are tailored to its architecture.

\subsection{Impact of augmented dataset size}
\begin{wraptable}{r}{0.5\textwidth}
    \centering
    \vspace{-4mm}
    \caption{Impact of training dataset size (from data augmentation) on task success rates.}
    \vspace{-2mm}
    \resizebox{\linewidth}{!}{
    \begin{tabular}{|c|c|c|c|c|}
        \cline{2-4}       
        \nocell{1} &  \multicolumn{3}{c|}{\textbf{Task Sucess Rate (\%)}} \\ \hline

        \textbf{Training Scheme} & {Bottle Collecting} & {Stacking Pringles} & {Hammering}\\ \hline
1k	& 53	&44	&37\\ \hline
5k	& 68	&55	&43\\ \hline
10k & 71	&62	&51\\ \hline
20k	& 83	&71	&57\\ \hline
40k	& 88	&77	&69\\ \hline
80k & \textbf{93}	&\textbf{84}	&\textbf{78}\\ \hline
120k 	&\textbf{93}	&\textbf{81}	&76\\ \hline
    \end{tabular}}
    \label{tab:dataset_size}
\end{wraptable}
To assess the limited diversity of trajectories derived from only a few demonstrations, we evaluated how policy performance varies with different amounts of augmented data. We trained policies on increasing subsets of the augmented data derived from the same 10 initial simulation demonstrations. Specifically, we trained policies with datasets of approximately $\{1000, 5000, 10000, 20000, 40000, 80000, 120000\}$ samples and evaluated them using 100 simulation experiments per task. 

For each subset, we sampled the trajectories uniformly from the full augmented dataset. We report the task success rate on the simulation experiments across three tasks: Bottle Collecting, Stacking, and Hammering. The results, presented in Table ~\ref{tab:dataset_size}, show how real-robot task success rates change as the number of augmented samples increases from 1k to 120k. For the bottle collecting task, the success rate improves substantially from 53\% with 1k samples to 93\% with 80k samples, with no further improvement observed at 120k, indicating a plateau in performance. In the stacking Pringles task, the success rate similarly improves from 44\% at 1k to 84\% at 80k, followed by a slight decline to 81\% at 120k, suggesting a saturation point or possible overfitting due to excessive data from a limited set of demonstrations. The hammering task, which involves more dynamic motion and tool use, shows the lowest initial performance at 37\% with 1k samples, but rises steadily to 78\% at 80k. A minor drop to 76\% at 120k again suggests that augmentation beyond a certain threshold may not be beneficial.
Overall, these results confirm that trajectory-level data augmentation is both effective and essential in scaling a small number of demonstrations into a dataset that supports robust policy learning. However, the improvements begin to saturate beyond approximately 80k augmented samples. This indicates that while our augmentation strategy successfully enriches the dataset’s diversity, there is a practical upper bound to the gains achievable from the given base demonstrations.

\subsection{Training data composition}
We trained several policies using different ratios of simulated to real-robot demonstrations, including $70\%-30\%$, $50\%-50\%$, $30\%-70\%$, and $0-100\%$ mixes. For each task, we recorded five real-world demonstrations and five simulation demonstrations, which were then augmented to create a larger dataset. The final training data was sampled from these augmented demonstrations according to the specified ratios. This approach allowed us to determine the optimal balance between simulation and real-world data for training robust and effective robot policies. To evaluate the performance of the robot on each task, we performed 10 real-robot experiments. Table~\ref{tab:data_combinition_results} presents the impact of varying the ratio of simulated to real-world demonstrations on policy performance across three manipulation tasks. 

\begin{wraptable}{r}{0.5\textwidth}
    \vspace{-4mm}
    \centering
    \caption{Task success rates (\%) for three real-robot tasks, Bottle Collecting, Stacking Pringles, and Hammering, trained using different ratios of simulated to real-robot demonstrations. Each training scheme used augmented data derived from five real and five simulated seed demonstrations, with the final training data sampled to reflect the specified ratio. Each reported value is based on 10 real-robot trials.}
    \vspace{2mm}
    \resizebox{\linewidth}{!}{
    \begin{tabular}{|c|c|c|c|c|}
        \cline{2-4}       
        \nocell{1} &  \multicolumn{3}{c|}{\textbf{Task Sucess Rate (\%)}} \\ \hline

        \textbf{Training Scheme} & {Bottle Collecting} & {Stacking Pringles} & {Hammering}\\ \hline
        Sim only (100-0) & 80 & 60 &50\\ \hline
        70-30 & \textbf{90} & \textbf{80} & \textbf{70}\\ \hline
        50-50 & 80 & 60  &50\\ \hline
        30-70 & 70 & 70 &60\\ \hline
        Real only (0-100) & 80 & 70 &60\\ \hline
    \end{tabular}}
    \label{tab:data_combinition_results}
\end{wraptable}

First, using simulated data only (100-0) results in reasonable baseline performance—80\% for Bottle Collecting, 60\% for Stacking Pringles, and 50\% for Hammering—demonstrating the general usefulness of simulation in training initial policies. However, the performance noticeably drops in tasks requiring fine-grained control, such as Hammering. This reflects the sim-to-real gap, where differences in object shape, contact interactions, and calibration errors between simulation and the real robot limit policy transferability. Introducing a small amount of real-world data helps bridge this gap. The 70\% simulated and 30\% real configuration yields the highest performance across all tasks. This demonstrates that even a limited injection of real demonstrations significantly improves the performance, while simulation continues to provide data diversity and scalability. In our experiments, this ratio effectively leverages the strengths of both domains. Interestingly, increasing the proportion of real data further (e.g., 50-50 or 30-70) does not yield further improvements. This result underscores that while real data mitigates sim-to-real issues, its limited diversity can constrain learning. Real demonstrations are time-consuming to collect and inherently less varied than thousands of augmented simulations, which better cover spatial and kinematic variations. Finally, real-only training (0-100) achieves results comparable to simulation-only for simpler tasks but underperformed in more complex tasks. This again highlights that neither domain is sufficient alone.


\section{Visualization of ground-truth and predicted trajectories }

\begin{figure}[!h]
    \centering  
    \includegraphics[width=\textwidth]{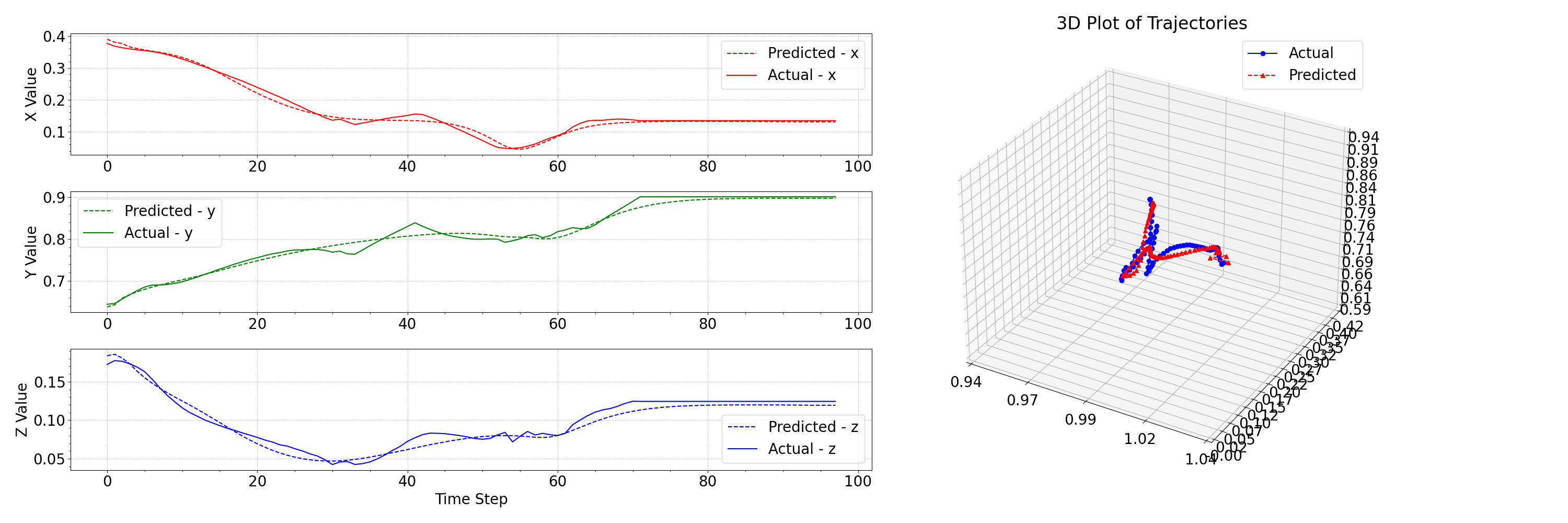}\\
    \includegraphics[width=\textwidth]{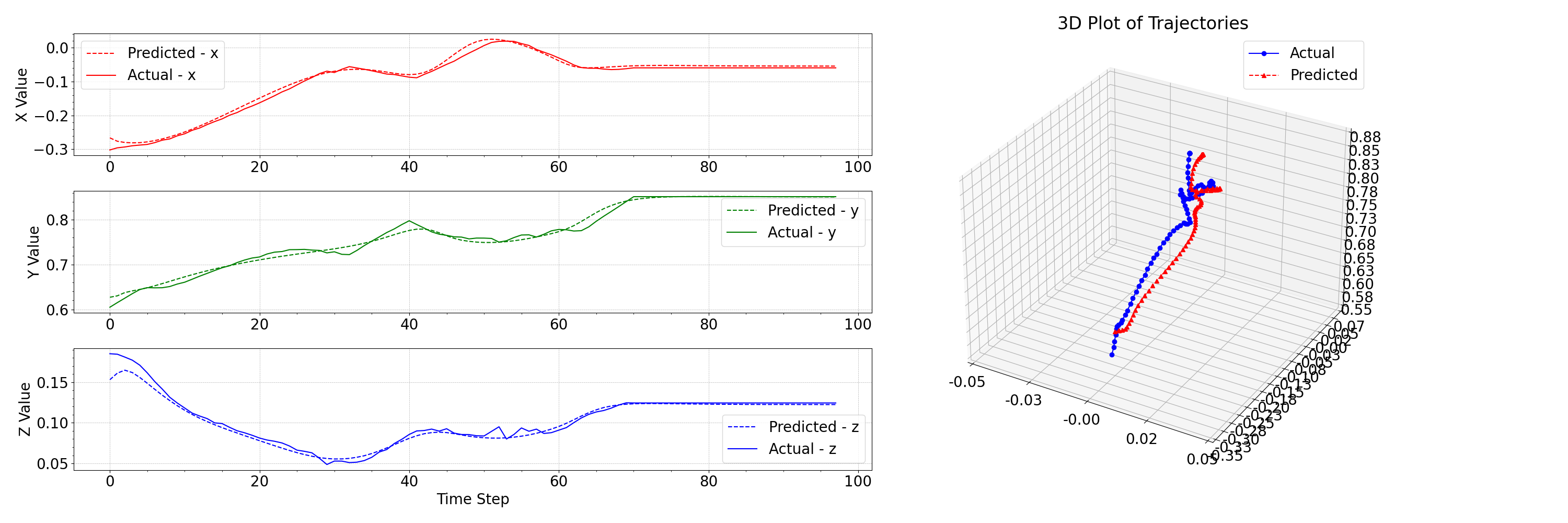}
    \caption{Comparative visualization of predicted and actual trajectories for two randomly selected samples. The plots illustrate the network's capability to accurately predict 98 intermediate waypoints for each sample. The 2D plots show the predicted versus actual values for the X, Y, and Z axes over time steps, while the 3D plots provide a comprehensive view of the trajectory paths. The close alignment between the predicted and actual trajectories in both the 2D and 3D visualizations highlights the accuracy and reliability of the network's predictions.  
    }
    \label{fig:lstm_prediction}
\end{figure}

\begin{figure}[!t]
    \centering  
    \includegraphics[width=\textwidth]{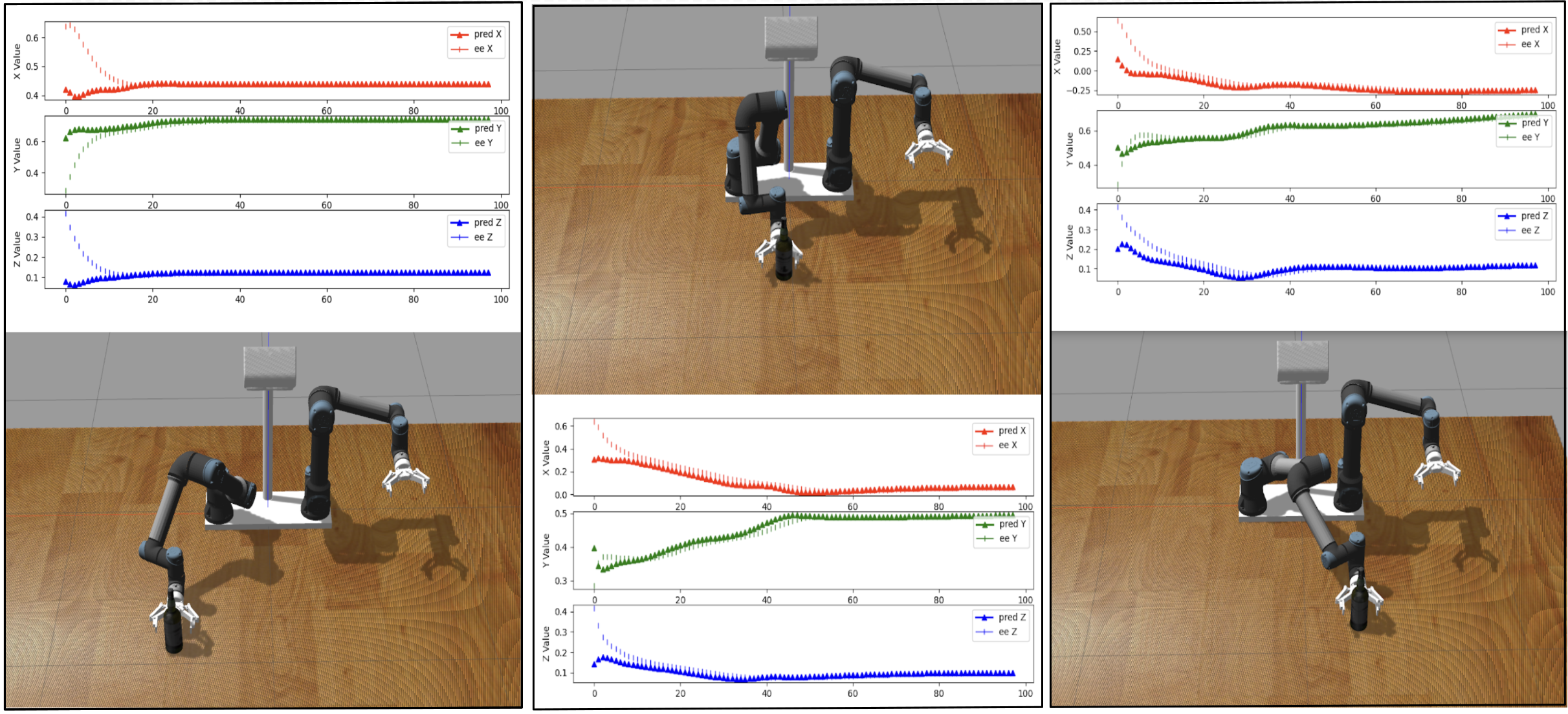}
    \caption{Grasping of a bottle object in various positions based on a single-shot demonstration (the first subtask of \textit{Bottle Collecting} task): The images and plots illustrate the robot's ability to grasp a bottle object in various poses after observing a single demonstration. The robot successfully follows the predicted trajectory to the pre-grasp area, as shown in the x, y, and z axes plots. These experiments demonstrate the proposed approach's effectiveness in enabling the robot to generalize and accurately execute grasping tasks in different scenarios. 
    }
    \vspace{-3mm}
    \label{fig:generalization}
\end{figure}

We learned a policy using a single-shot demonstration and tested its performance on the validation set of the first subtask of the bottle-collecting task. Specifically, we plot 2D/3D visualization of predicted and actual trajectories for two randomly selected samples from the validation set. The network's performance is depicted through both 2D and 3D plots in Figure~\ref{fig:lstm_prediction}. Furthermore, we tested the performance of the policy on the digital twin of the robot by randomly placing a bottle object in front of the robot, retrieving the current pose of the end effector and the pose of the object, and forwarding them as input to the learned policy. We then executed the predicted trajectory and visualized the robot's performance. Figure~\ref{fig:generalization} shows the performance of the robot in three experiments, highlighting its ability to follow the predicted trajectory and accurately grasp the object based on the learned policy. The visualizations include both images of the robotic arm and plots of the predicted versus actual trajectories for the X, Y, and Z axes.

\clearpage
\section {Experiments}
\begin{wrapfigure}{r}{0.35\textwidth}
\vspace{-5mm}
 \includegraphics[width=0.35\textwidth]{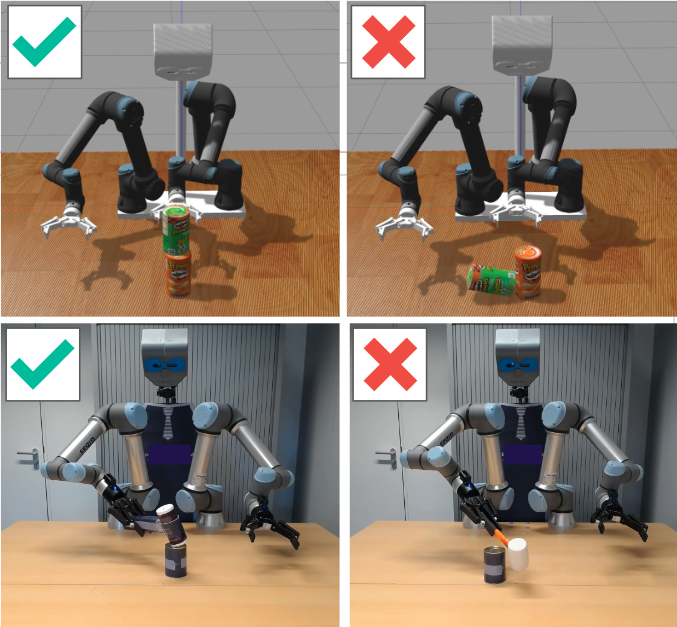}
    \caption{Successes and failures for stacking and hammering tasks.}
    \label{fig:sucess_and_failure}
    \vspace{-5mm}
\end{wrapfigure}
\subsection{Success and failure criteria for the proposed tasks}
For the collecting objects task, success is determined by placing all objects into the basket and failure is considered if one of the objects is not placed into the basket and remains outside. Figure~\ref{fig:sucess_and_failure} shows examples of successful (left) and failed (right) attempts for the stacking and hammering tasks. In the stacking task, success is achieved when objects are stacked stably. In the hammering task, the success criteria are based on hitting the target object accurately.  These criteria help in evaluating the performance and reliability of the robotic system in executing these tasks. In the case of setting up a drink tray task, success is considered when the robot places all items (such as a mug and a bottle) accurately on their designated spots on the tray, and failure is defined when items are not placed correctly on the tray, or the robot fails to place all items.

\begin{figure}[!b]
    \centering  
    \vspace{-5mm}
    \includegraphics[width=0.8\textwidth]{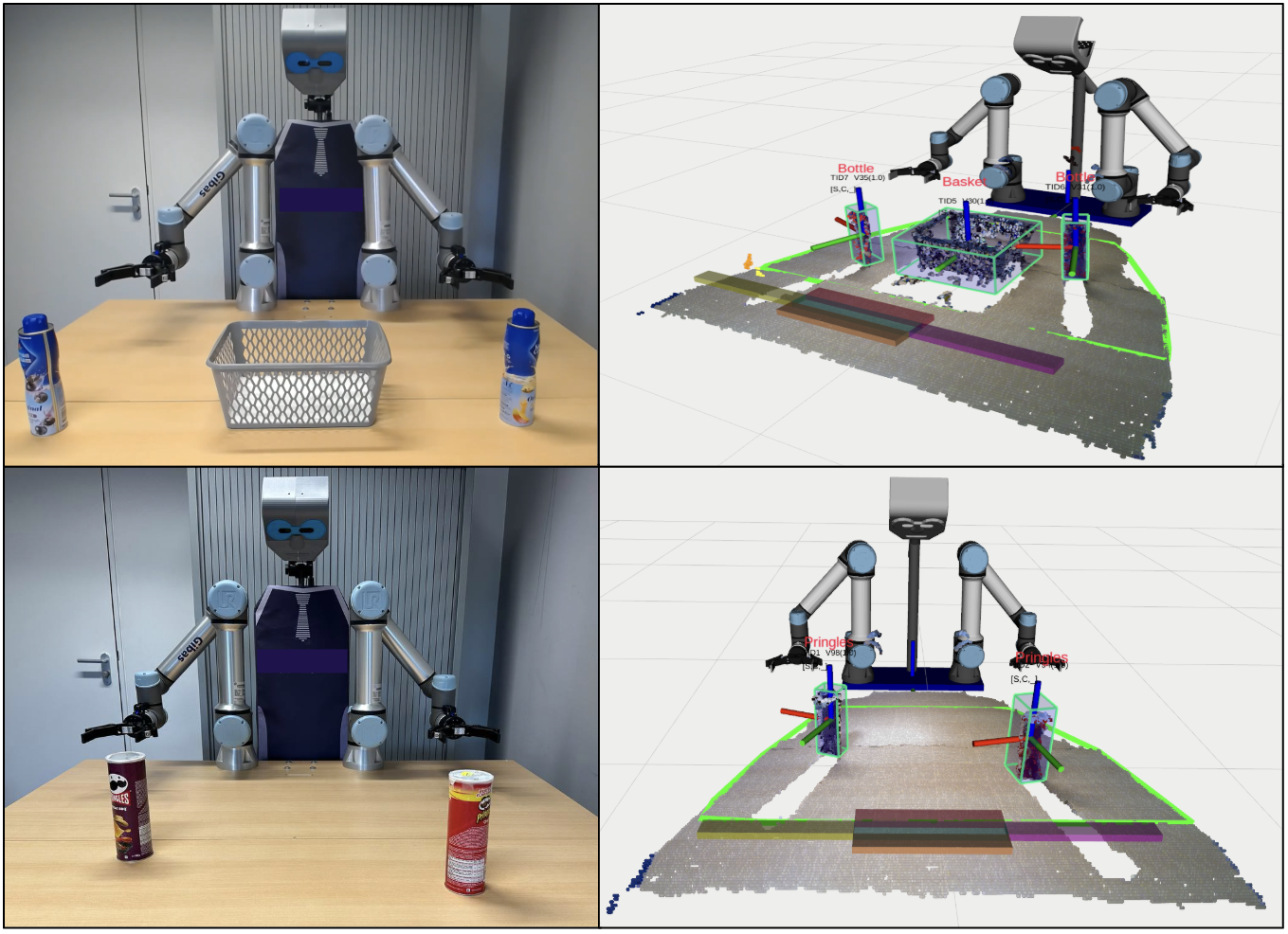}
    \caption{Our real robot setup. The \textit{left} column shows the physical hardware, including a dual-arm configuration composed of two Universal Robots (UR5e), each equipped with a Robotiq 2F-140 gripper, and an Asus Xtion RGB-D camera mounted for perception. The \textit{right} column visualizes the robot’s perception system in action, capturing the workspace through the RGB-D sensor. Detected objects are annotated with semantic labels, 3D bounding boxes, and estimated poses. These object poses, along with the current end-effector state, are used as inputs to the learned policy, which predicts the trajectory to execute at each step of the task. }
        \label{fig:collecting_bottles_real_robot}
\end{figure}
\subsection{Perception in Real-Robot Experiments}

To support robust policy execution in the real world, we rely on an integrated perception system that enables the robot to sense and interpret its environment. Figure~\ref{fig:collecting_bottles_real_robot} illustrates our experimental setup for the bottle collecting and stacking tasks, highlighting both the physical hardware and the robot’s visual understanding of the scene. The robot is equipped with an Asus Xtion RGB-D camera, which provides real-time depth and color data. Using this input, the perception module performs object detection, tracking, and recognition. As shown in Fig.~\ref{fig:collecting_bottles_real_robot}, all objects in the scene are localized and recognized within the robot’s workspace. 

In our approach, this perceptual information is essential for closing the loop between sensing and action. The estimated object poses, combined with the robot's current end-effector configuration, are used as inputs to the learned policy. At each step, the policy consumes these observations and predicts the next trajectory segment required to complete the task. This enables the robot to adapt to variations in object positioning and ensures reliable execution even under changes in the environment. The results demonstrate that the perception system provides accurate and timely information, allowing the robot to perform complex manipulation tasks with high success rates.


\subsection{Effectiveness of Human-in-the-Loop Corrections on Real Robot}

As discussed in Section 4.2-Q3 in the main paper, we observed that fine-tuning the policy using human-in-the-loop (HITL) corrections improved task performance. After incorporating corrections from failed real-robot experiments and fine-tuning the policy, we noted an average improvement of approximately 5\% in simulation. To evaluate whether these improvements translated to real-world performance, we conducted 20 new trials per task, 10 using the original policy and 10 using the fine-tuned policy. To ensure a fair comparison, we carefully maintained consistent object configurations across both sets of trials. Specifically, for each object arrangement, we first ran an experiment using the original policy, then reset the scene to closely match the previous configuration before executing the same task with the fine-tuned policy.

The results demonstrated a clear performance improvement on the real robot. For the bottle collecting task, the success rate remained consistent at 90\% for both policies, as this task is relatively simple and less sensitive to control variations. However, for the stacking task, the success rate increased from 80\% to 90\%, and for the hammering task, from 70\% to 90\%. These results confirm the value of incorporating real-world correction signals into training. The fine-tuned policy is better adapted to real-world uncertainties such as object pose variations, actuation delays, and perception noise, which are often not fully captured in simulation.

\subsection{Generalization Performance}
We evaluated the generalization performance of our learned policies across three scenarios to assess robustness and adaptability. Specifically, we considered: (i) spatial generalization with known tasks and known objects, (ii) object generalization with known tasks and new objects, and (iii) task generalization with entirely new tasks and new objects. These evaluations provide insights into how well the policy transfers to variations in object placement, object properties, and task structure without additional fine-tuning.

\subsubsection{Spatial generalization with known tasks and known objects}
\begin{figure}[!t]
    \centering  
    \includegraphics[width=\textwidth]{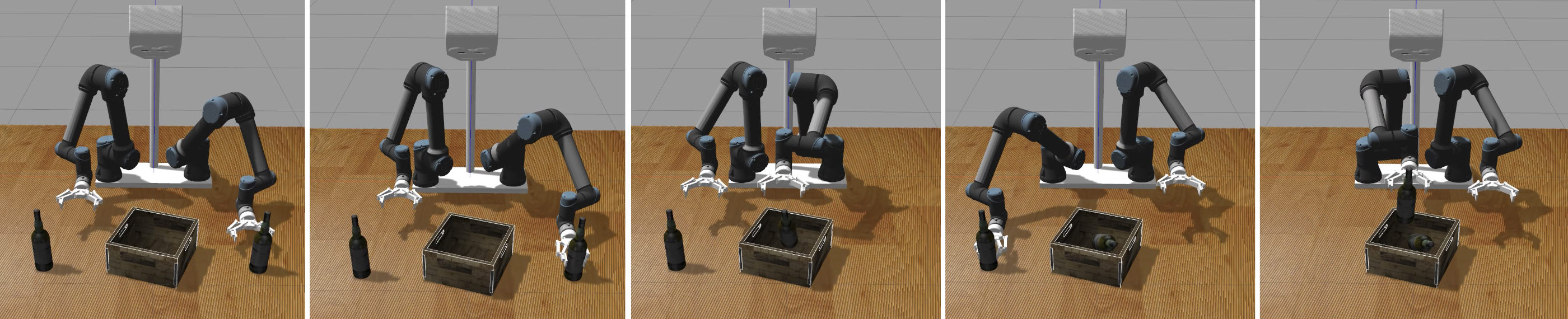}\\
    \includegraphics[width=\textwidth]{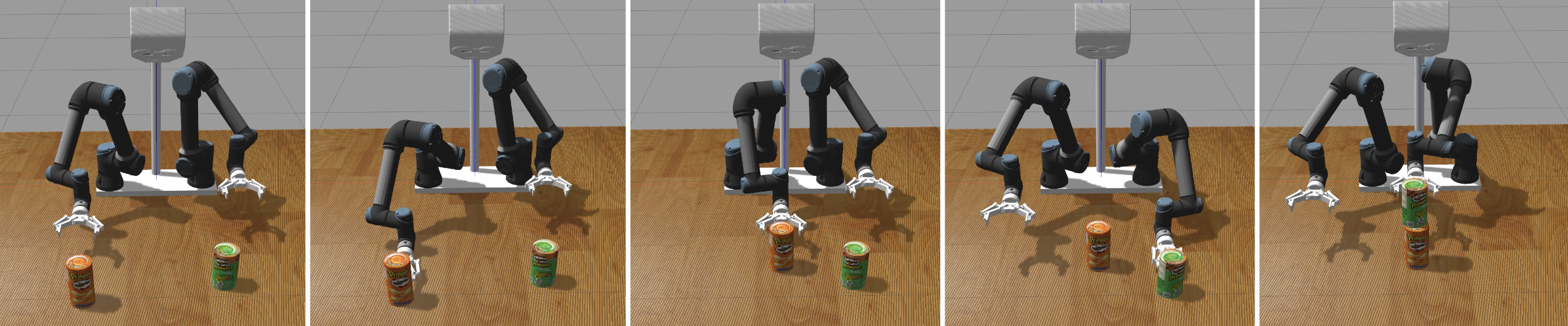}\\
    \includegraphics[width=\textwidth]{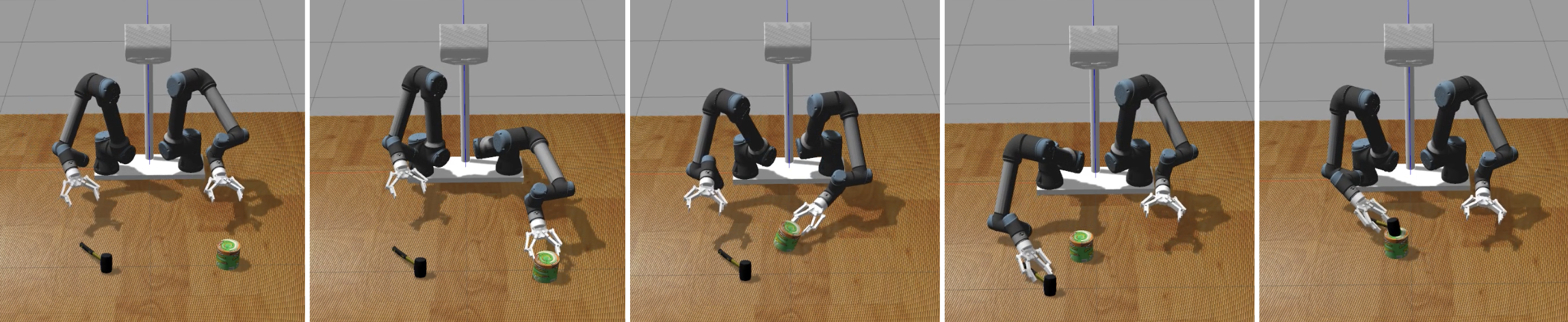}\\
    \includegraphics[width=\textwidth, trim=2mm 2mm 2mm 2mm, clip=true]{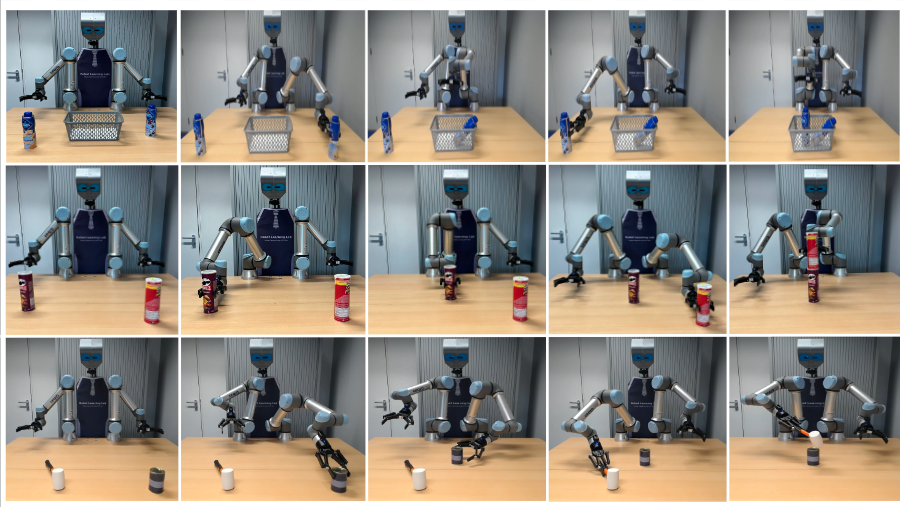}

    \caption{Step-by-step execution of three tasks by the robot in both simulation and the real world. The top three rows show the robot performing (top row) bottle collecting, (second row) stacking objects, and (third row) hammering in simulation. The bottom three rows depict the same tasks performed by the real robot in a physical environment. This sequence demonstrates the robot's ability to predict a sequence of subtasks and follow a predicted trajectory to complete the task successfully.
    }
    \label{fig:all_tasks}
\end{figure}

In this experiment, we tested the robot’s ability to generalize spatially by varying the initial positions of the objects by up to $\pm20~cm$ along the $x$ and $y$ axes. This setup maintains both the object identities and task types but introduces variation in spatial configuration.
Figure~\ref{fig:all_tasks} illustrates the robot's performance across three tasks, collecting bottles, stacking objects, and hammering, each executed using the same objects as in the training demonstrations. Despite the perturbations in object placement, the robot successfully completed the tasks by accurately following the predicted trajectories. These results highlight the spatial robustness and reliability of the learned policies. Quantitatively, in simulation, the robot achieved a success rate of 92\% in the bottle collecting task, 85\% in the stacking task, and 86\% in the hammering task. These high success rates demonstrate strong spatial generalization within known task and object domains.  To evaluate the same generalization behavior on the real robot, we repeated each task for five trials on the real robot under similar spatial variations. The robot achieved success rates of 4/5 in bottle collecting, 3/5 in stacking, and 4/5 in hammering. These results reflect the real-world challenges such as perception noise, and actuation delays.



\newpage
\subsubsection{Object generalization with known tasks and new objects}

In this experiment, we evaluate the robot’s ability to generalize to previously unseen objects while performing familiar tasks. Specifically, we replace the original objects with five new items that vary in weight, size, and shape, including water bottles (250\,ml and 1\,L), a Pringles can, a Coke can, a juice bottle, baskets, hammers, and stackable bottles. These objects differ from those used during training demonstrations. This evaluation aims to assess the policy’s robustness when exposed to variations in object geometry and dynamics, without altering the overall task structure or scene configuration. Despite the changes in physical properties, the robot successfully follows the predicted trajectories and completes the object collection task without requiring additional fine-tuning.

\begin{figure}[!h]
    \centering  
    \includegraphics[width=\textwidth]{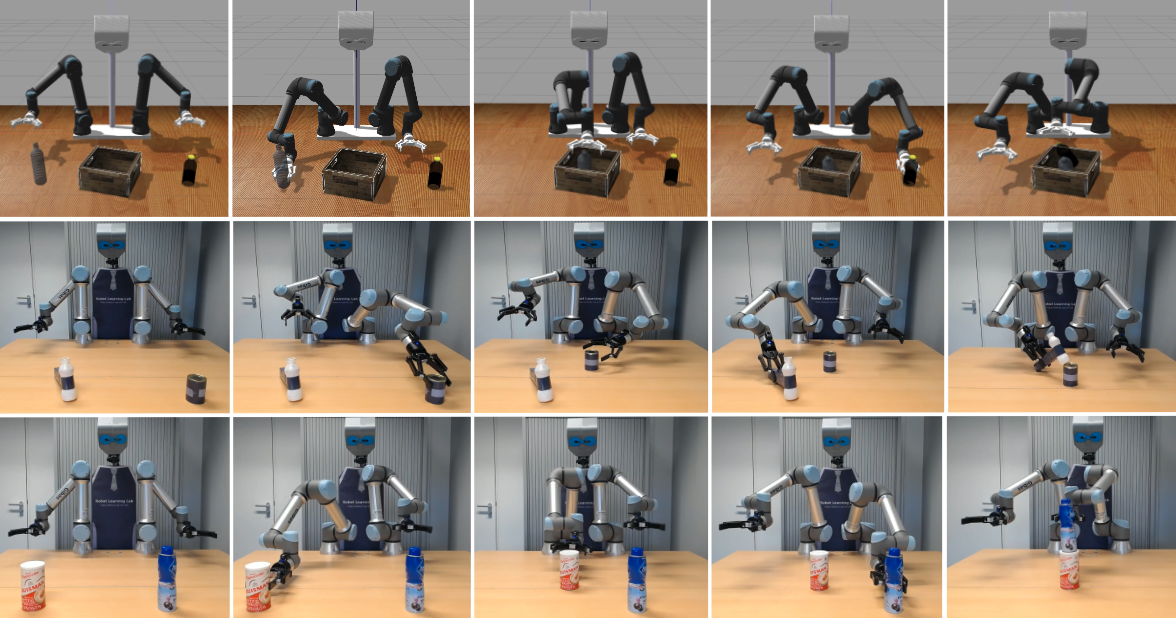}\\
    \caption{ Evaluation of new object generalization in both simulation (top row) and real-robot experiments (middle and bottom rows). The robot successfully executes the bottle collecting and stacking tasks with new objects placed in perturbed initial positions. Real-robot experiments (middle: hammering; bottom: stacking bottles) demonstrate successful task execution, highlighting the learned policy’s robustness and ability to generalize beyond training conditions.
    }
    \vspace{-3mm}
    \label{fig:collecting_new_bottles}
\end{figure}

To quantify performance, we conducted 100 trials in simulation for each task using the new object set, and five trials in real-robot settings. The robot achieved a success rate of 95\% in simulation and 4/5 on the real robot for the object collection task. For stacking with new bottle types, the success rates were 87\% in simulation and 3/5 in real-world trials. These results indicate a slight performance drop due to sim-to-real discrepancies and physical variability, but overall, the policy demonstrates good generalization to new object instances. Figure~\ref{fig:collecting_new_bottles} illustrates a sequence of snapshots showing the robot’s performance with the new objects. The results demonstrate that the learned policy is capable of generalizing across object variations, highlighting its adaptability and robustness in real-world applications.

\subsubsection{Task generalization with entirely new tasks and new objects}

We further evaluated the generalization capabilities of the learned policies by testing the robot in completely new scenarios with unfamiliar objects. Specifically, we introduced two new tasks: (i) setting up a drink tray with a cup and a bottle placed on the tray, and (ii) stacking two plastic cups. These tasks were not part of the training distribution, and the objects used differed in appearance, size, and weight from those seen during training. To perform these tasks, we reused the stacking policy and modified the end poses of the subtasks to suit the new objectives. No retraining or additional supervision was applied, highlighting the flexibility of the learned policy in handling zero-shot task variations.

For each task, we conducted 100 trials in simulation and 10 trials in real-robot settings. In the drink tray setup, the robot achieved a success rate of 92\% in simulation and 8/10 on the real robot. For the cup stacking task, performed only on the real robot, the policy achieved a success rate of 87\% in simulation and 7/10, with failures primarily due to minor misalignments during placement. These results demonstrate that the policy can generalize beyond its training domain, handling novel task objectives and object geometries with only minimal task-specific adjustments.

\begin{figure}[!h]
    \centering  
    \includegraphics[width=\textwidth]{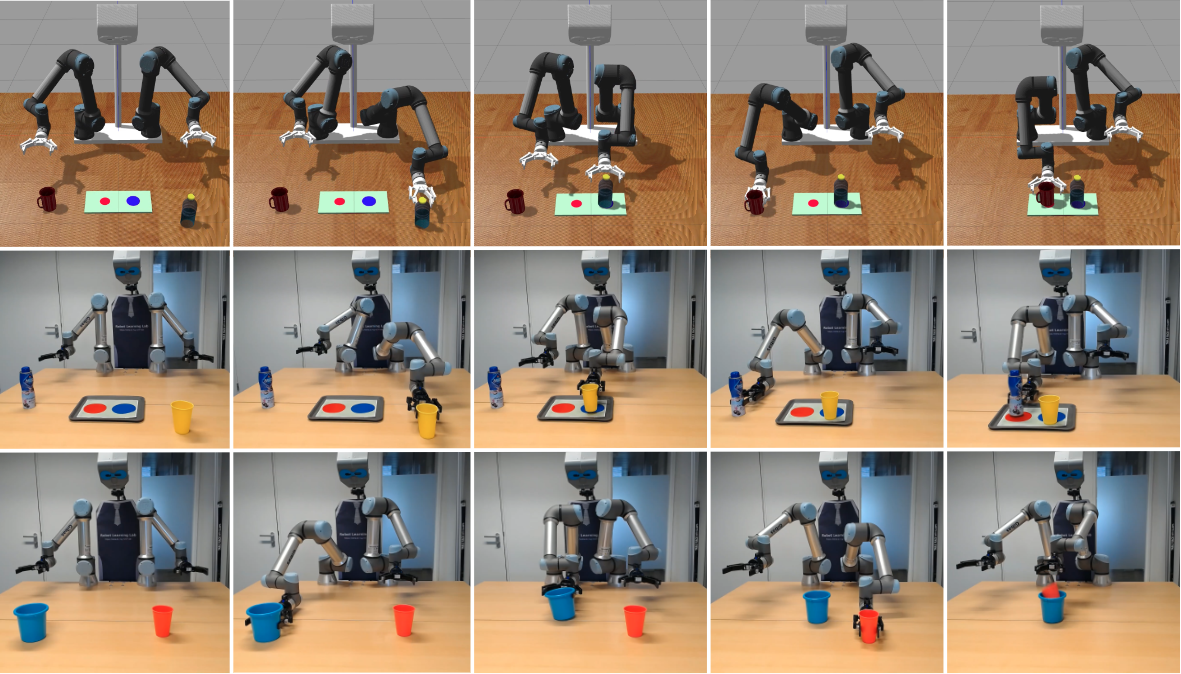}
    \caption{The images illustrate the robot's step-by-step execution of the task, demonstrating its ability to adapt to new scenarios and handle new objects. 
    The sequence captures the robot's interaction with various items and its precise placement of each item on the tray, showcasing the generalization and adaptability of the robotic system. 
    }
    \label{fig:new_scenarios_new_objects}
\end{figure}

Figure~\ref{fig:new_scenarios_new_objects} shows the sequence of snapshots detailing the robot's performance in setting up a drink tray (first two rows in simulation and real robot), and stacking cups (last row in real robot). This image highlights the robot's capacity to generalize from its training and successfully perform tasks in new scenarios with new objects.

\begin{figure}[!h]
    \centering  
    \includegraphics[width=\textwidth]{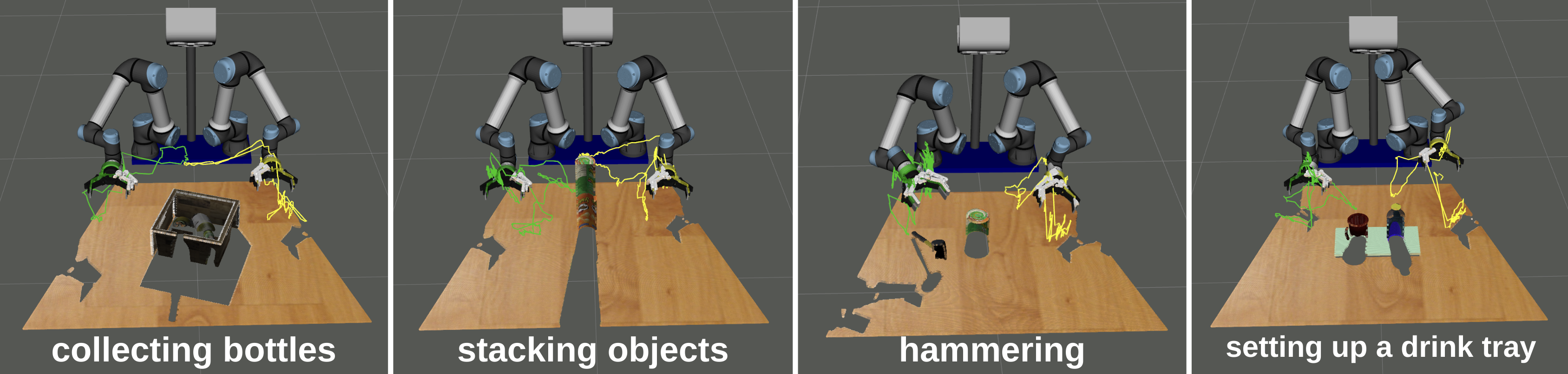}
    \caption{The highlighted predicted trajectories represent the planned paths that the robot follows to complete each task: collecting bottles, stacking objects, hammering, and setting up a drink tray.}
    \label{fig:trajectories_visualziation}
\end{figure}

For the tasks shown in Figure~\ref{fig:trajectories_visualziation}, we visualize the predicted trajectories used to guide the robot’s motion. The trajectories are color-coded, with yellow indicating the path of the left arm and green for the right arm. Each trajectory line corresponds to a sequence of waypoints generated by our policy, which are conditioned on the current state of the robot and the estimated poses of the relevant objects. This visual demonstration underscores how the learned policy integrates perception and control to generate reliable, task-specific motion plans in diverse scenarios.


\section {Failure Cases During Real Robot Experiments}

\begin{figure}[!t]
    \centering  
    \includegraphics[width=\textwidth]{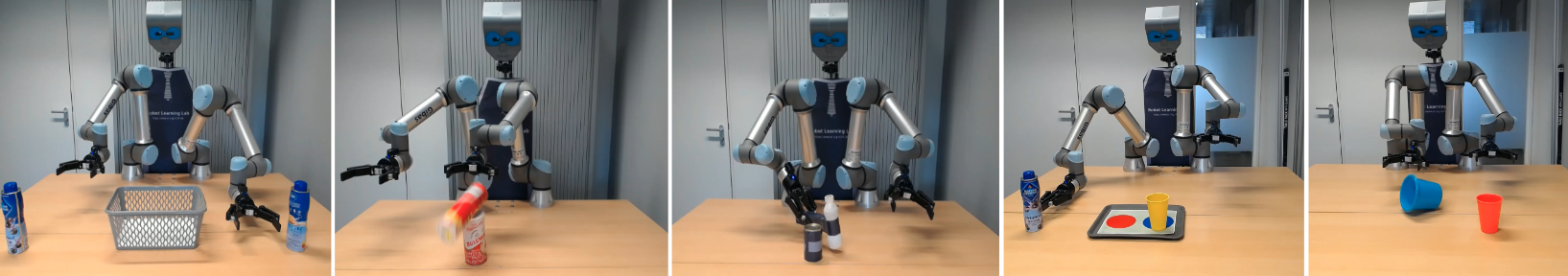}
    \caption{ Various failure cases observed during the real robot experiments. From left to right: (1) The robot fails to accurately grasp and place the bottles into the basket, likely due to errors in trajectory prediction. (2) In the stacking task, the robot is unable to align the objects stably, resulting in toppling or misplacement; this could happen due to the sim-to-real gap or inaccurate object pose estimation. (3) During hammering, the robot fails to strike the target accurately, potentially due to open-loop control limitations and differences in object shape or position. (4) In the drink tray task, the robot failed to grasp the bottle. (5) In the cup stacking task, the bottom or the top cup is either misaligned or falls off, likely due to unstable placement. }
    \label{fig:failure_cases}
\end{figure}

During the real robot experiments, several failure cases were observed that can be attributed to the differences between simulation and real-world conditions (See Fig.~\ref{fig:failure_cases}). These failures provide valuable insights into the challenges of transferring learned policies from simulation to real robots. The main factors contributing to these failures include:

\textbf{Gap between simulation and real robot:} The objects used during the experiments in the real world differ in shape, size, weight, and material properties compared to those used in the simulation. These discrepancies can affect the robot's ability to accurately predict and execute the required trajectories, leading to unsuccessful task completion.

\textbf{Inverse kinematics solver and command frequency:} The inverse kinematics (IK) solver and the frequency of sending commands to the robot differ between the simulation and the real robot. In the simulation, the IK solver and command frequency are typically more idealized and may operate at higher rates, leading to smoother motions. In contrast, the real robot's controller might introduce slight delays due to various factors (e.g., network loads), causing deviations in the robot's movement and impacting task execution.

\textbf{Trajectory prediction errors:} Inaccuracies in the predicted trajectory can lead to the robot's end-effector being misaligned with the object, preventing successful grasping and manipulation. Poor grasping or inadequate grip strength can cause objects to slip during manipulation. 

\textbf{Execution of actions:} Using an open-loop controller during the execution of actions means that the robot follows the predicted trajectories blindly without feedback corrections. In real-world scenarios, this lack of feedback can lead to accumulated errors and deviations from the intended path, causing failures in task completion. In such an open-loop scenario, accurate calibration between the robot's camera and its manipulator is crucial and calibration errors can result in incorrect positioning and unsuccessful task execution.

These factors highlight the importance of addressing the sim-to-real gap and improving the robustness of robotic systems when transitioning from simulation to real-world applications. To mitigate these challenges and improve reliability, future work should explore the integration of closed-loop feedback controllers that predict the next few waypoints based on the current observation, rather than generating the entire trajectory upfront. 

\end{document}